\documentclass[twoside,11pt]{article}
\usepackage{jmlr2e}
\usepackage{amsmath,amssymb}
\newcommand{\sign}{\text{sign}}
\usepackage{todonotes}

\ShortHeadings{Complex-Valued Gaussian Process Regression}{Ambrogioni et al}

\newcommand*\conj[1]{\bar{#1}}

\usepackage{lineno}
\begin{document}
\vspace*{0.35in}

\begin{flushleft}
{\Large
\textbf\newline{Complex-valued Gaussian Process Regression for Time Series Analysis}
}
\newline
\\
Luca Ambrogioni\textsuperscript{1} and
Eric Maris\textsuperscript{1}
\\
\bigskip
\bf{1} Radboud University, Donders Institute for Brain, Cognition and Behaviour, Nijmegen, The Netherlands 
\\
\bigskip
* e.maris@psych.ru.nl \\
* l.ambrogioni@donders.ru.nl

\end{flushleft}

\begin{abstract}
The construction of synthetic complex-valued signals from real-valued observations is an important step in many time series analysis techniques. The most widely used approach is based on the Hilbert transform, which maps the real-valued signal into its quadrature component. In this paper, we define a probabilistic generalization of this approach. We model the observable real-valued signal as the real part of a latent complex-valued Gaussian process. In order to obtain the appropriate statistical relationship between its real and imaginary parts, we define two new classes of complex-valued covariance functions. Through an analysis of simulated chirplets and stochastic oscillations, we show that the resulting Gaussian process complex-valued signal provides a better estimate of the instantaneous amplitude and frequency than the established approaches. Furthermore, the complex-valued Gaussian process regression allows to incorporate prior information about the structure in signal and noise and thereby to tailor the analysis to the features of the signal. As a example, we analyze the non-stationary dynamics of brain oscillations in the alpha band, as measured using magneto-encephalography. 
\end{abstract}


\section{Introduction}
In this paper, we introduce the use of complex-valued Gaussian process (GP) regression for the construction of complex-valued signals. The principal aim of this construction is to quantify the instantaneous amplitude, phase and frequency of real-valued oscillatory signals. GP regression is a powerful tool for the analysis of time series which relies on a prior model of their temporal autocorrelations, as specified by a covariance function \citep{rasmussen2006gaussian, brahim2004gaussian, reece2010introduction}. We define two new families of complex-valued covariance functions that induce cross-correlation between the real and the imaginary part of complex-valued signals. This statistical relation is then exploited for constructing the (unobservable) imaginary part from the (observable) real part of the signal. The resulting complex-valued GP regression (CGPR) is a very flexible tool that can be adapted to a wide range of signals by choosing a particular covariance function that is tailored to the specific properties of the signal. 

In the rest of the paper, we introduce the basic concepts of CGPR and its relevance for time series analysis. Our main goal is to improve the estimation of the instantaneous frequency and amplitude of real-valued signals. The most commonly used method for instantaneous frequency and amplitude estimation involves the use of the Hilbert transform to construct the complex-valued analytic representation of real-valued signals \citep{boashash1992estimatingI}. Therefore, we start our exposition by reviewing the construction of the analytic representation. Then, after a short review of the GP regression framework, we introduce the CGPR and the construction of the associated complex-valued signal. The properties of the resulting complex-valued signal depend on the choice of the covariance function. We introduce two new classes of covariance functions that reproduce the properties of the analytic representation in expectation. We denote these classes of functions as \emph{quadrature} and \emph{quasi-quadrature} covariance functions. 

In a series a simulation studies, we validate our method on several classes of oscillatory signals, both in the noise-free and in the noisy case. We compare the performance of our method with the analytic representation (based on the Hilbert transform) and the Morlet wavelet analysis. These two methods are still the most used in real-word applications and have strong theoretical connection with our new method. Finally, we apply our method on neural data measured using magnetoencephalography (MEG).

\subsection{Related works}
In the recent years, several authors extended the application of kernel methods to complex-valued data. In a seminal paper, Bouboulis and Theodoridis introduced both the use of the complex Gaussian kernel and the general technique of kernel complexification, which is used to obtain a complex-valued kernel from an arbitrary real-valued kernel \citep{bouboulis2011extension}. This work has been subsequently extended to non-circular complex-valued data, i.e. data that are sampled from distributions that are not invariant under rotations of the complex plane \citep{bouboulis2012augmented, tobar2012novel}. Using hermitian covariance functions, \citet{boloix2014gaussian} recently extended GP regression to complex-valued signals. This formulation of the complex-valued GP regression was then extended to non-circular complex-valued signals \citep{boloix2015complex}. 

The main aim of these cited publications is to apply kernel methods to the analysis of complex-valued signals. This differs from our current approach, in which we use complex-valued GP regression to analyze real-valued signals. To this aim, we introduced two new families of Hermitian covariance functions and a different observation model. In this paper, we will consider stationary GP priors. The Fourier transform of a stationary GP has uniformly distributed phases. As a consequence, the so-called pseudo-covariance (relation) matrix is zero, and the complex normal distribution becomes circularly-symmetric. This restriction to stationary GPs implies, amongst others, that our prior stochastic process cannot have a fixed time relation to the onset of the signal \citep{schreier2003stochastic}. 

In spite of its well-known shortcomings, the analytic representation, based on the Hilbert transform, is still considered to be an important benchmark method for the quantification of instantaneous amplitude, phase and frequency of real-valued signals \citep{girolami2002instantaneous,le2001comparison, turner2011probabilistic}. A major limitation of this approach is that it cannot be directly applied to noise-corrupted signals. This has motivated the development of Bayesian methods that incorporate an explicit noise model. For example, the probabilistic phase vocoder (PPV) is based on a discrete-time first-order Gaussian autoregressive process whose latent variables (i.e. the real and the imaginary part) can be estimated using the Kalman smoother algorithm \citep{cemgil2005probabilistic}. Since these processes are Gaussian, this approach can be seen as a simple special case of our GP regression method. Recently, \citet{turner2011probabilistic} improved the PPV method by specifying a prior distribution for the phase and amplitude processes. This leads to a more complex non-conjugate Bayesian model whose posterior needs to be approximated using a computationally intensive expectation propagation scheme. Conversely, our GGPR, like the PPV method, can be solved in closed-form. CGPR also differs from the existing Bayesian models in the fact that our prior processes are continuous-time. Thus, our model conceptually separates the sampling process (as determined by the likelihood) from the underlying signal (as determined by the GP prior).

Several real-wold signals are mixtures of narrow-band components. The analytic representation approach produces meaningful quantifications of instantaneous amplitude, phase and frequency only when the original real-valued signal is narrow-band. Therefore, mixed signals must be decomposed into their narrow-band components prior to the application of the Hilbert transform. A possible approach is to use a family of wavelets, each centered at the mean frequency of a narrow-band component \citep{gao1999instantaneous}. This approach has a strong theoretical connection with our method. In fact, as we shall show, Morlet wavelet analysis can be seen as a special case of CGPR. An improvement on the wavelet approach is the temporal GP decomposition, which allows for an estimation of the covariance of each component, while at the same time explicitly modelling the broad-band noise \citep{ambrogioni2016dynamic}. An increasingly popular decomposition approach is the Hilbert–Huang transform (HHT), which is based on a combination of empirical mode decomposition and the Hilbert transform \citep{huang2014hilbert}. This method is particularly suitable for non-linear and non-stationary signals. Because CGPR is an alternative for the Hilbert transform, it can also be applied to the components obtained using empirical mode decomposition.

\section{Theoretical background}
In this section, we will review some theoretical results that are relevant for the understanding of the rest of the paper. In particular, we will introduce the quadrature filter (and the related Hilbert transform) as a tool for constructing the complex-valued analytic representation of a real-valued signal which in turn will be used to quantify instantaneous amplitude, phase and frequency. Finally, we will briefly describe the Morlet wavelet method and its relations with the analytic representation. 

\subsection{Instantaneous amplitude and phase} Given a complex-valued signal $z(t)$, its instantaneous amplitude and phase are defined through its polar representation:
\begin{equation}
z(t) = |z(t)| \exp{\big(i\arg[z(t)]\big)} \qquad,
\label{polar representation, methods}
\end{equation}
in which the complex modulus $|z(t)| = \sqrt{ (\mathfrak{R}z(t))^2 + (\mathfrak{I}z(t))^2}$ is called instantaneous amplitude and $\arg[z(t)] = -i \log \frac{z(t)}{|z(t)|}$ is called instantaneous phase. Furthermore, the instantaneous (angular) frequency $\omega(t)$ is defined as the first derivative of the instantaneous phase
\begin{equation}
\int_0^t \omega(s) ds = \arg[z(t)] - \arg[z(0)].
\label{instantaneous frequency, methods}
\end{equation}
In order to extend these definitions to real-valued signals, it is natural to replace the complex exponential in Eq. \ref{polar representation, methods} with a real-valued sinusoid:
\begin{equation}
s(t) = \mathfrak{A}(t) \cos \mathfrak{P}(t)~,
\label{polar representation (real), methods}
\end{equation}
where $\mathfrak{A}(t)$ is a positive-valued amplitude function and $\mathfrak{P}(t)$ is a phase function. Unfortunately there are infinitely many ways of representing an arbitrary real-valued signal in this form. For example, any real-valued signal can be written as $$s(t) = |s(t)|\cos{\big(\pi/2 - \pi/2 ~\sign(s(t))\big)}$$ but the functions $|s(t)|$ and $\big(\pi/2 - \pi/2 ~\sign(s(t))\big)$ cannot be meaningfully interpreted as respectively instantaneous amplitude and phase. Intuitively, the functions in the right hand side of Eq.~\ref{polar representation (real), methods} can be interpreted as instantaneous amplitude and phase when the rate of change of $\mathfrak{A}(t)$ is much smaller than the rate of change of the oscillation $\cos \mathfrak{P}(t)$. In the next subsection, we will review the most well-known formalization of this intuitive idea.

\subsection{Analysis of a simple family of band-limited signals} We will now see how to construct the instantaneous amplitude and phase for a simple family of real-valued signals. In doing so, we will follow a general strategy that will also be useful in the case of arbitrary band-limited signals. The idea is to define a transform that converts the original signal into a complex-valued signal whose instantaneous amplitude and phase can be extracted using Eq.~\ref{polar representation (real), methods}. Consider the following deterministic signal
\begin{equation}
s(t) = \mathfrak{A}(t) \cos{(2 \pi f_0 t)}~,
\label{test signal, methods}
\end{equation}
where $\mathfrak{A}(t)$ is an arbitrary positive-valued function that has most of its energy at frequencies lower than $f_0$. It is intuitive to consider $s(t)$ as an oscillation with instantaneous amplitude $\mathfrak{A}(t)$ and instantaneous phase $2 \pi f_0 t $. Hence, we aim to construct a complex-valued signal $z(t)$ such that $|z(t)| = \mathfrak{A}(t)$ and $\arg[z(t)] =  2 \pi f_0 t$. This can be done by applying the formula $\cos(a) = \frac{1}{2} \big(\exp{(i a)} + \exp{(-i a)} \big)$ from which follows that $s(t)$ decomposes in the sum of a counterclockwise part $s_{+}(t)$ and a clockwise part $s_{-}(t)$:
\begin{equation}
s(t) = \frac{1}{2} \big(s_{+}(t) + s_{-}(t)\big)
\label{positive and negative frequencies, methods}
\end{equation}
where
\begin{equation}
s_{+}(t) = \mathfrak{A}(t) \exp{\big(+ i 2 \pi f_0 t\big)}~, ~~s_{-}(t) = \mathfrak{A}(t) \exp{\big(- i 2 \pi f_0 t\big)}~,
\end{equation}
We can now show that the complex-valued signal $z(t)$ constructed from $s(t)$ should be its counterclockwise part. In fact:
\begin{equation}
|s_{+}(t)| = \mathfrak{A}(t)~
\end{equation}
and
\begin{equation}
\arg[s_{+}(t)] = 2 \pi f_0 t~.
\end{equation}

\subsection{Quadrature filter and analytic representation} Most real-world signals do not have a simple analytic expression such as Eq.~\ref{test signal, methods}. Nevertheless, it is possible to devise general purpose transforms which provide an intuitively plausible quantification of instantaneous amplitude and phase. The most commonly used transform of this type is the quadrature filter \citep{boashash1992estimating}. The action of the quadrature filter $\mathcal{A}$ on a real-valued signal $s(t)$ is defined by the following formula
\begin{equation}
\mathcal{A} s(t) = s(t) + i \mathcal{H} s(t), 
\label{analytic signal, methods}
\end{equation}
where
\begin{equation}
\mathcal{H} s(t) = \frac{1}{\pi}\int_{-\infty}^{+\infty} \frac {s(\tau)}{t - \tau} d\tau
\label{hilbert transform, methods}
\end{equation}
is called the Hilbert transform of $s(t)$. Here and in the following, expressions such as $\mathcal{H} s(t)$ denote the application of the linear operator $\mathcal{H}$ on the signal $s(t)$. The filter kernel of the Hilbert transform is the function $\frac{1}{\pi t}$ and, by taking its formal Fourier transform, we obtain the frequency response function of the Hilbert transform:
\begin{equation}
\tilde{\mathcal{H}}(\xi) = -i ~\textrm{sgn}(\xi) = \exp{\big(-i ~\textrm{sgn}(\xi) \frac{\pi}{2}\big)}.
\label{hilbert transform fourier, methods}
\end{equation}
Hence, the Hilbert transform acts on the signal by phase-shifting the Fourier coefficients for positive and negative frequencies by, respectively, $-\pi/2$ and $\pi/2$. In the rest of the paper, we will refer to this relation between the Fourier coefficients of the real and the imaginary part as the \textit{quadrature relationship}. 
Using Eq.~\ref{hilbert transform fourier, methods}, it can be shown that the frequency response function of the quadrature filter is
\begin{equation}
\tilde{\mathcal{A}}(\xi) = \big(1 -  i^2 \textrm{sgn}(\xi)\big)  = 2 h(\xi) 
\label{quadrature frequency response, methods}
\end{equation}
where $h(\xi)$ is the Heaviside step function. This means that the transformed signal $\mathcal{A} s(t)$ is always analytic, meaning that it has all its Fourier coefficients associated with the negative frequencies equal to zero. Therefore, the quadrature filter maps a signal into its unique analytic representation, defined as the complex-valued signal that has its real part identical to the original signal while having all the Fourier coefficients associated with the negative frequencies equal to zero \citep{boashash1992estimating}. This mapping does not entail any information loss since, for real-valued signals, the Fourier coefficients of the negative frequencies are the complex conjugate of their positive frequency counterparts. Summarizing, the quadrature filter is a linear operator while the analytic representation is the complex-valued signal resulting from the application of the filter on a real-valued signal.

\subsection{Bedrosian's theorem} Bedrosian's theorem \citep{wang2009simple} shows that the quadrature filter can perfectly separate the envelope function from the oscillatory part when their spectra are disjoint. More formally, consider a signal of the form given by Eq.~\ref{polar representation (real), methods}. If there exists a positive number $a$ such that the spectrum of $\mathfrak{A}(t)$ is identically equal to zero for $|\xi|>a$ while the spectrum of $\cos\mathfrak{P}(t)$ is identically equal to zero for $|\xi|<a$, then we have that
\begin{equation}
\mathcal{A} [\mathfrak{A}(t) \cos\mathfrak{P}(t)] = \mathfrak{A}(t) \mathcal{A} [\cos\mathfrak{P}(t)] = \mathfrak{A}(t) \exp{\big(i \mathfrak{P}(t)\big)}~.
\label{bedrosian theorem, methods}
\end{equation}
This theorem justifies the use of the analytic representation for the quantification of the instantaneous amplitude and phase of a real-valued signal. 

\begin{figure}[!ht]
    	\includegraphics[width=1.\textwidth] {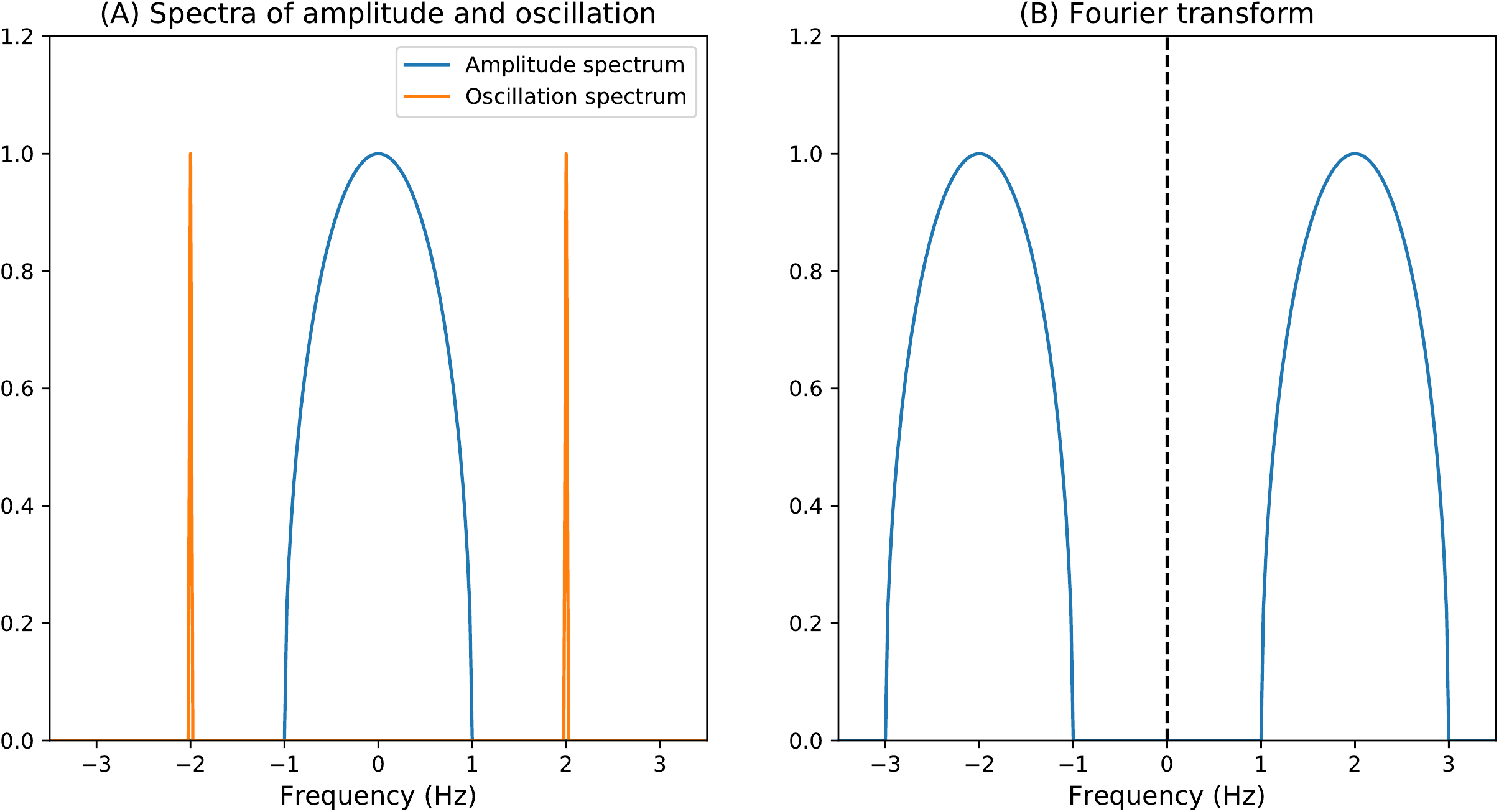}
	\caption{Visualization of Bedrosian Theorem on the example signal. A) Spectrum of the envelope (blue) and oscillation (green). B) Signal in the frequency domain, the counterclockwise (clockwise) part has energy only in the positive (negative) side of the spectrum.}
	\label{figure 1}
\end{figure}

In the following, we will give an intuitive proof of Bedrosian's theorem for the special case of a signal of the form given by Eq.~\ref{test signal, methods}. From the previous analysis we know that, in order to correctly recover $\mathfrak{A}(t)$ and $\mathfrak{P}(t)$, the complex-valued signal constructed from $s(t)$ should be its counterclockwise part $s_{+}(t)$. Therefore, we need to show that, when the conditions of Bedrosian's theorem are met, $\mathcal{A}s(t)$ is equal to $s_{+}(t)$. This can be done by taking the Fourier transform of $s(t)$. Using the convolution theorem, together with the fact that the Fourier transform of the complex exponential $\exp{(i y t)}$ is the delta function $\delta(\xi - y)$, it can be shown that this Fourier transform is
\begin{equation}
\tilde{s}(\xi) = \frac{1}{2} \big( \tilde{s}_{+}(\xi) + \tilde{s}_{-}(\xi) \big)~,
\label{fourier test signal, methods}
\end{equation}
where
\begin{equation}
\tilde{s}_{+}(\xi) = \tilde{\mathfrak{A}}(\xi - 2 \pi f_0)~,~~ \tilde{s}_{-}(\xi) = \tilde{\mathfrak{A}}(\xi + 2 \pi f_0)
~,
\label{fourier clockwise frequency test signal, methods}
\end{equation}
are the Fourier transforms of the counterclockwise part and of the clockwise part respectively (see Eq.~\ref{positive and negative frequencies, methods}). Recall that, in the frequency domain, the analytic representation $\tilde{\mathcal{A}s}(\xi)$ 
is obtained by setting to zero all the Fourier coefficients associated with the negative frequencies and multiplying by two. Hence, $\tilde{\mathcal{A}s}(\xi)$ is identically equal to $\tilde{s}_{+}(\xi) = \tilde{\mathfrak{A}}(\xi - 2 \pi f_0)$ if and only if $\tilde{\mathfrak{A}}(\xi)$ vanishes when $|\xi|>2\pi f_0$. In fact, in this case, the counterclockwise and the clockwise parts of the signal have energy only at the positive and negative side of the spectrum respectively and, consequently, they can be perfectly separated by the quadrature filter. Figure \ref{figure 1} shows the spectral separation of the counterclockwise and clockwise parts for an example signal (in the frequency domain). 

\subsection{Limitations of the quadrature filter} 
\begin{figure}[!ht]
	\centering
    	\includegraphics[width=1.\textwidth] {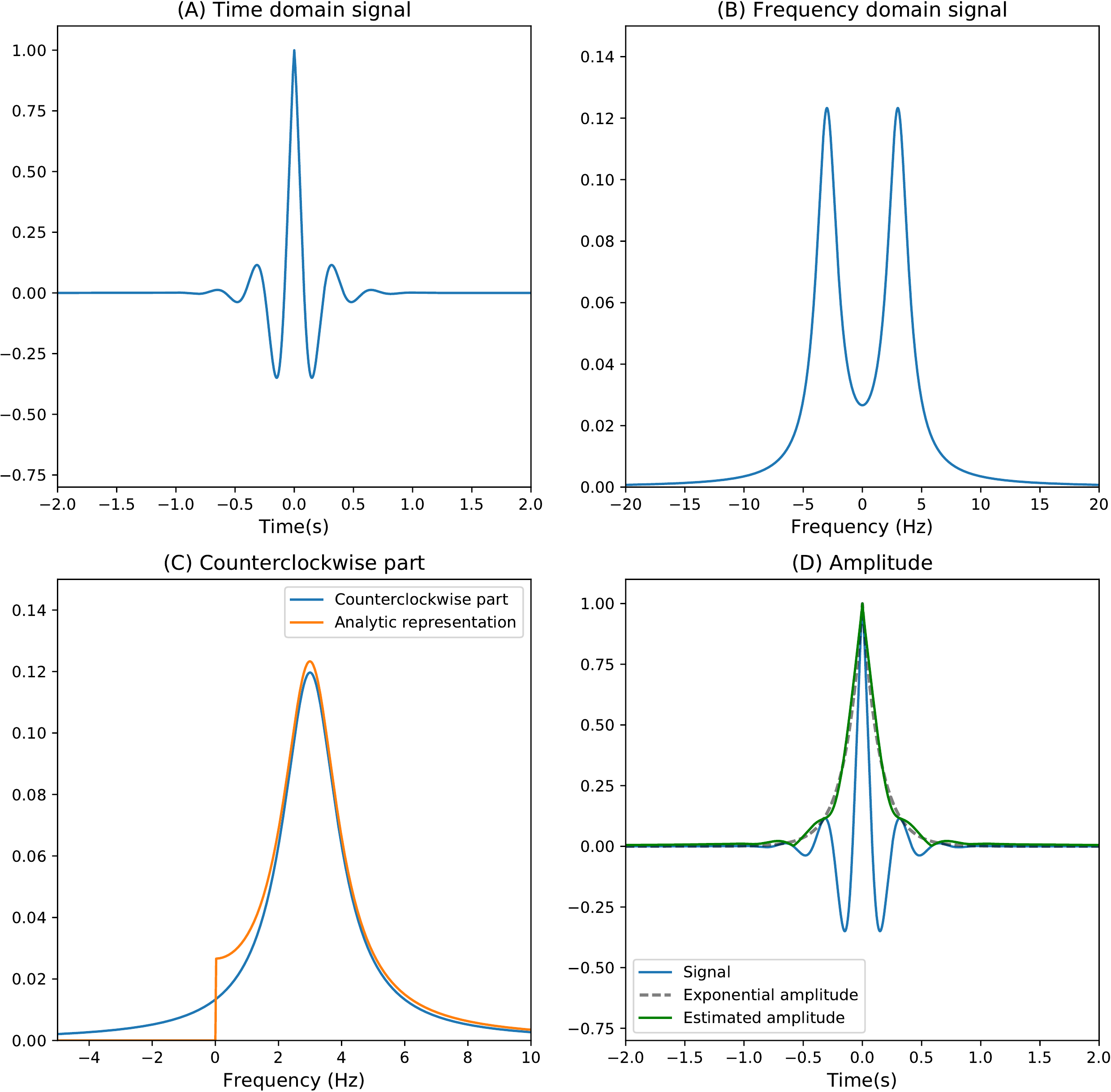}
	\caption{An example signal. A) Time course of the example signal. B) Fourier transform of the example signal. C) Clockwise part (blue) and analytic representation (green) in the frequency domain. D) Signal envelope (dashed) and analytic construction of the envelope (green).}
	\label{figure 2}
\end{figure}

Bedrosian's theorem is important for understanding the limitations of the quadrature filter. In particular, the condition of disjoint spectra is too restrictive for most real life signals and can lead to rather counterintuitive estimates of the instantaneous amplitude and phase. For example, consider the signal $s(t) = \exp{(-\frac{|t|}{\sigma})}\cos{(2\pi f_0 t)}$ where $\sigma$ is a positive constant. This signal (with $\sigma = 0.15$ and $f_0 = 3$) is plotted in Fig.~\ref{figure 2}A and its spectrum is plotted in Fig.~\ref{figure 2}B. Importantly, the spectra of $\exp{(-\frac{|t|}{\sigma})}$ and $\cos{(2\pi f_0 t)}$ are not disjoint because the Fourier transform of $\exp{(-\frac{|t|}{\sigma})}$ is different from zero for all frequencies. This implies that the counterclockwise part has energy at negative frequencies and it is not equal to the analytic representation of the signal (see Fig.~\ref{figure 2}C). Consequently, the analytic representation does not recover $\mathfrak{A}(t)$ and $\mathfrak{P}(t)$ exactly. However, the spectral overlap is very small if $\sigma \gg 1/f_0$ and, under this condition, the estimates $|\mathcal{A}s(t)|$ and $\arg[\mathcal{A}s(t)]$ are very close to, respectively, $\mathfrak{A}(t)$ and $\mathfrak{P}(t)$. Conversely when $\sigma \approx 1/f_0$, the estimates can differ dramatically from $\mathfrak{A}(t)$ and $\mathfrak{P}(t)$. Fig.~\ref{figure 2}D shows the amplitude constructed using the analytic representation ($\sigma = 0.15, f_0 = 3$). Note that this function is not unimodal and has oscillating tails. More dramatic examples of this kind of behavior can be found in the Results section.

\subsection{Morlet wavelet analysis}
In this subsection, we will review the Morlet wavelet method for phase estimation \citep{gao1999instantaneous}. This approach is used in many applications, especially when the measured signal is noisy or is a mixture of several narrow-band components \citep{le2001comparison}. 

A Morlet wavelet is obtained by multiplying a complex harmonic signal with a Gaussian envelope:
\begin{equation}
W(t) = \frac{1}{\sqrt{\pi \mu^2}} \exp{\big(-\frac{t^2}{2 \mu^2} + i \omega_0 t\big)}~,
\label{morlet wavelet, methods}
\end{equation}
where $\omega_0$ is the center frequency and $\mu$ determines its temporal localization. 

We can construct a complex-valued signal from the real-valued continuous-time signal $s(t)$ by convolving it with the complex-conjugate of the wavelet:
\begin{equation}
s_w(t) = \int_{-\infty}^{+\infty}  s(\tau) W^{*}(\tau - t) d \tau~.
\label{wavelet convolution, methods}
\end{equation}
The effect of this transformation is clearer in the frequency domain. Using the Fourier convolution theorem on Eq.~\ref{wavelet convolution, methods}, we obtain:
\begin{equation}
s_w(\xi) = W(\xi) s(\xi) = \frac{1}{\sqrt{\pi}}\exp{\big(-\mu^2 \frac{(\xi - \omega_0)^2}{2}\big)} s(\xi)~.
\label{wavelet frequency response, methods}
\end{equation}
Hence, the convolution with the Morlet wavelet can be interpreted as a band-pass filter with a Gaussion frequency response function.

Importantly, when $\omega_0 \gg \mu$, the energy of the filter is almost completely concentrated on the positive frequency semi-axis and, consequently, the resulting complex-valued signal is approximately analytic. Therefore, to the extent that the energy of $s(\xi)$ is concentrated around $\omega_0$, it is meaningful to quantify the instantaneous phase of $s(t)$ as the phase of the complex-valued signal $s_w(t)$. 

It is important to stress that the real part of $s_w(t)$ is in general different from $s(t)$ since the wavelet method involves a band-pass filter operation. This is different from the quadrature operator which does not affect the real part. The filtering operation has the advantage of converting any arbitrary $s(t)$ into a band-limited signal whose instantaneous phase can be quantified in a meaningful way. Since most real-wold signals are corrupted by measurement noise, in most applications, a filtering operation is required. Consequently, some kind of band-pass filter is usually applied prior to the quadrature operator. Note however that $s_w(t)$ is not identical to the analytic representation of a band-pass filtered version of $s(t)$ since $s_w(\xi)$ is not identically equal to zero for negative values of $\xi$. As we noted in the previous subsection, this deviation from analyticity can improve the estimation of instantaneous amplitude and phase. Unfortunately, in the Morlet wavelet method both the level of noise reduction and of deviation from analyticity are regulated by the single localization parameter $\mu$. 

This is not optimal since, as we showed in the previous subsection, the exact analyticity of the complex-signal can be problematic even when the underlying real-valued signal is noise-free and band limited. In the following sections, we will show that the Morlet wavelet method is a special case of a newly introduced family of methods where the amount of denoising and of deviation from analyticity can be regulated independently.

\section{Gaussian process regression}
We now review the basic concepts of real-valued GP regression, laying the groundwork for the definition of our new families of GP covariance functions. We will begin by briefly reviewing standard real-valued GP regression. The complex-valued extension easily follows from the real-valued case and will allow us to obtain a construction of the complex-valued signal in a probabilistic framework.

\subsection{Real-valued Gaussian Process regression}
A GP generalizes the multivariate Gaussian distribution to infinitely many degrees of freedom. More formally, a stochastic process $\psi(t)$ follows a GP distribution when all its marginal distributions $p(\psi(t_1),...,\psi(t_n))$ are multivariate Gaussians \citep{rasmussen2006gaussian}. Since these Gaussians can be parametrized in terms of their mean and covariance matrices, a GP is fully specified by its mean and covariance functions, respectively
\begin{equation}
m_\psi(t) = \langle \psi(t) \rangle
\label{mean function, methods}
\end{equation}
and
\begin{equation}
k_\psi(t_1,t_2) = \langle (\psi(t_1) - m_\psi(t_1)) (\psi(t_2) - m_\psi(t_2)) \rangle.
\label{covariance function, methods}
\end{equation}
In order to properly specify a GP distribution, $k_\psi(t_1,t_2)$ has to be symmetric and positive semi-definite, meaning that all the covariance matrices of the marginal distributions are symmetric and positive semi-definite.

In the remaining, we assume the mean function $m_\psi(t)$ to be zero for all values of $t$. Furthermore, we restrict our attention to stationary covariance functions, i.e. functions that solely depend on the time lag $k(t_1, t_2) = k(t_2 - t_1) = k(\tau)$. In this case, we can define the \textit{spectral density} (or \textit{power spectrum}) $\tilde{k}(\xi)$ as the Fourier transform of $k(\tau)$ 
\begin{equation}
\tilde{k}(\xi) = \frac{1}{2 \pi}\int_{-\infty}^{+\infty} \exp{\big(-i \xi \tau \big)} k(\tau) d\tau.
\label{spectral density, methods}
\end{equation}
The usefulness of GPs in machine learning follows from the fact that GP distributions can be used as priors for a non-parametric Bayesian regression. In particular, given a set of $n$ sample points $T = \{t_1,....t_n\}$, we can model a sampled signal $s_j$ as follows:
\begin{equation}
s_j = \psi(t_j) + \epsilon_j
\label{observation model GP, methods}
\end{equation}
where the vector $\boldsymbol{\epsilon}$, whose $j$-th entry is $\epsilon_j$, is a multivariate noise term that we assume to be Gaussian with a (potentially non-diagonal) covariance matrix $Q$. Note that the sample points do not need to be equally spaced, although equal spacing can be convenient for numerical reasons.

The aim of this regression problem is to estimate the latent function $\psi(t)$ without resorting to a specific parametric form (e.g. linear, polynomial, sinusoidal, wavelet). Using GP regression, we can estimate the value of $\psi(t)$ at any arbitrary set of $m$ target points $T^{\times} = \{t^{\times}_1,...,t^{\times}_m \}$, even if they are not part of our sample points. To this end, it is convenient to define the vector $\boldsymbol{\psi}$ with entries $\psi_j = \psi(t^{\times}_j)$. Since both the prior distribution of $\boldsymbol{\psi}$ and the likelihood $p(s|\boldsymbol{\psi})$ are multivariate Gaussians, the posterior distribution $p(\boldsymbol{\psi}|s)$ is also a multivariate Gaussian. In particular, its expected value $\boldsymbol{m}_{\psi|s}$ is given by:
\begin{equation}
\boldsymbol{m}_{\psi|s} = K^{\times}_\psi (K_\psi + Q)^{-1} \boldsymbol{s},
\label{posterior expectation GP, methods}
\end{equation}
where the sample covariance matrix $K_\psi$ is defined by the entries $[K_\psi]_{jk} = k_\psi(t_j,t_k)$ and the sample-to-target cross-covariance matrix $K^{\times}_\psi$ by the entries $[K^{\times}_\psi]_{jk} = k_\psi(t^{\times}_j,t_k)$ \citep{rasmussen2006gaussian}. From this formula we can see that the matrix $K^{\times}_\psi$ projects the information from the sample points to the target points by leveraging our prior model of the temporal correlations determined by the function $k_\psi(t_1,t_2)$.

A similar formula applies if we aim to reconstruct a completely unobservable new process $\rho(t)$ from the observation of $\psi(t)$ once we assume their prior cross-covariance function $c_{\rho,\psi}(t_1,t_2) = \langle \rho(t_1) \psi(t_2) \rangle$. This point is crucial for our current goals as it will allow us to reconstruct the unobservable imaginary part from a measured real-valued signal. If we restrict our attention to the sampled time points, the posterior expectation of $\boldsymbol{\rho}$ given $\boldsymbol{s}$ is
\begin{equation}
\boldsymbol{m}_{\rho|s} = C_{\rho,\psi} (K_\psi + Q)^{-1} \boldsymbol{s},
\label{cross posterior expectation GP, methods}
\end{equation}
where the $j,k$-th entry of the cross-covariance matrix $C_{\rho,\psi}$ is $c_{\rho,\psi}(t_j,t_k)$. 
We conclude this subsection by discussing the computational complexity of GP regression. The computational bottleneck of Eq.~\ref{posterior expectation GP, methods} and Eq.~\ref{cross posterior expectation GP, methods} is the inversion of the covariance matrix. The inversion of a positive definite symmetric matrix is usually performed using the Cholesky decomposition. This decomposition of the covariance matrix has a cubic complexity with respect to the number of time points \citep{krishnamoorthy2011matrix}. 
 
\subsection{Complex-valued Gaussian Process regression}
We can now generalize GP regression to the case in which the latent process $\zeta(t) = \alpha(t) + i \beta(t)$ is complex-valued. In order to construct a complex-valued GP we assume all the marginals of $\zeta(t)$ to be circularly-symmetric complex normal distributions \citep{picinbono1996second}. The resulting stochastic process is fully specified by the Hermitian, positive semi-definite function $k_\zeta(t_1, t_2) =  \langle \zeta(t_1) \conj{\zeta}(t_2) \rangle$, where $\conj{\zeta}(t_2)$ is the complex conjugate of $\zeta(t_2)$. The positive semi-definite Hermitian function $k(t,t')$ generalizes the concept of covariance function to complex-valued GPs. All the eigenvalues of Hermitian matrices are real. Moreover, Hermitian matrices have orthogonal eigenvectors for distinct eigenvalues \citep{arfken2005mathematical}. The logarithm of the probability density of a circularly-symmetric complex-valued Gaussian vector $v$ with Hermitian covariance $K$ is (up to a normalization constant) equal to $-v^{*}K^{-1}v$ \citep{picinbono1996second}. Given the similarity of this expression with the real-valued case, $K$ is often simply denoted as covariance matrix. In analogy with the real-valued case, we refer to $k_\zeta(t_1, t_2)$ as the covariance function of the process. 

In the remaining of the subsection we will derive the posterior expectation of the complex-valued signal $\zeta(t)$ for an arbitrary Hermitian covariance function $k_\zeta(t_1, t_2)$. We must assume that only the real part of $\zeta(t)$ generates $\boldsymbol{s}$ whereas the imaginary part is not observable
\begin{equation}
s_j = \mathfrak{R} \zeta(t_j) + \epsilon_j = \alpha(t_j) + \epsilon_j~.
\label{complex observation model GP, methods}
\end{equation}
Here, we assume that the measurements are corrupted by a noise term $\epsilon_j$. This is an important difference with the usual construction of the analytic representation, which cannot account for measurement noise. 

The posterior expectation of $\zeta(t_j)$ can be obtained using Eq.~\ref{cross posterior expectation GP, methods}. In fact, since the complex distribution is circularly symmetric, the prior cross-covariance matrix between the real and the imaginary part at the sample points is given by $\mathfrak{I} K_\zeta$, which denotes the entrywise imaginary part of the matrix $K_\zeta$ \citep{picinbono1996second}. Analogously, the prior covariance matrix of both the real and the imaginary part is $\mathfrak{R} K_\zeta$, i.e. the entrywise real part of $K_\zeta$. Thus, from Eq.~\ref{cross posterior expectation GP, methods}, we can see that the posterior expectation of the complex-valued signal $\zeta(t)$ at the sample points is 
\begin{equation}
\boldsymbol{m}_{\zeta|s} = (\mathfrak{R}K_\zeta + i \mathfrak{I} K_\zeta)(\mathfrak{R} K_\zeta + Q)^{-1} \boldsymbol{s} = K_\zeta(\mathfrak{R} K_\zeta + Q)^{-1}  \boldsymbol{s},
\label{complex posterior expectation GP, methods}
\end{equation}

\section{Quadrature and quasi-quadrature covariance functions}
Our construction of a GP-based complex signal depends on the choice of the prior covariance function. The aim of this section is to introduce a class of covariance functions that are suitable for a probabilistic generalization of the analytic representation. In the following, we will introduce two families of covariance functions with slightly different properties. As we shall see, quadrature covariance functions enforce strict analyticity of the resulting complex-valued GP and can be used for a probabilistic reformulation of the analytic representation. Quasi-quadrature covariance functions soften the analyticity constraint and give rise to a method that has higher performance when the spectra of instantaneous amplitude and phase components are highly overlapping. We begin by defining quadrature and quasi-quadrature covariance functions as the result of specific linear operators which map real-valued stationary covariance functions into complex-valued covariance functions. We then show that GPs equipped with those complex-valued covariance functions respect some important properties. Specifically, we define the concept of quadrature relationship in expectation as a generalization of the deterministic quadrature relationship induced by the quadrature filter. In subsection 4.1, we show that the real and the imaginary part of quadrature GPs have a deterministic quadrature relationship. In subsection 4.2, we then show that the real and the imaginary part of quasi-quadrature GPs have a quadrature relationship in expectation and that this relationship approaches an exact quadrature relationship asymptotically for frequency going to infinity.

\subsection{Quadrature covariance functions}
One of the defining features of the analytic representation is that, for any given frequency, the Fourier coefficients of the imaginary part are obtained by a $\pm \frac {\pi} {2}$ phase shift of the coefficients of the real part (see Eq.~\ref{hilbert transform fourier, methods}). Our aim is to reinterpret this relation in a probabilistic sense.  In particular, we say that the process $\zeta(t) = \alpha(t) + i \beta(t)$ is in \textit{quadrature relationship in expectation} when
\begin{equation}
{\langle \tilde{\beta}(\xi) \rangle}_{\tilde{\alpha}(\xi)} = c(\xi) \exp{\big(-i ~\textrm{sgn}(\xi) \frac{\pi}{2} \big)} \tilde{\alpha}(\xi) 
\label{quadrature relation in expectation, methods}
\end{equation}
where $\tilde{\alpha}(\xi)$ and $\tilde{\beta}(\xi)$ are the Fourier transforms of, respectively, the real and the imaginary part, and $\langle \tilde{\beta}(\xi) \rangle_{\tilde{\alpha}(\xi)}$ denotes the conditional expectation of $\tilde{\beta}(\xi)$ given $\tilde{\alpha}(\xi)$. In this expression, the positive scalar $c(\xi)$ ranges from zero to one and represent the strength of the statistical association between the Fourier coefficients of the real and the imaginary part. 

\subsubsection{Definition and general results}
In order to construct a GP that is in quadrature relationship in expectation, we define the quadrature covariance function induced by an arbitrary (real-valued) stationary covariance function $k(\tau)$ as follows:
\begin{equation}
\mathcal{A} k(\tau) = k(\tau) + i \mathcal{H} k(\tau).
\label{quadrature covariance function, methods}
\end{equation}
This formula defines a valid covariance function, i.e. a function that is Hermitian and positive semi-definite (see appendix I). We will now show that the Fourier coefficients of any GP defined by a covariance function of this form are indeed quadrature related in expectation. To this end, it is convenient to organize the real and the imaginary part of the process $\zeta(t)$ into the real-valued vector process
\begin{equation}
\Psi(t) = 
\begin{pmatrix}
    \alpha(t)\\
    \beta(t)
\end{pmatrix}.
\label{vector process, methods}
\end{equation}
Since $\zeta(t)$ is governed by a circularly-symmetric GP with a quadrature covariance function, the (matrix valued) \textit{cross-covariance function} of $\Psi(t)$ is given by the following formula (see appendix II for a derivation)
\begin{equation}
\Theta(\tau) = \langle \Psi(t)\Psi(t + \tau)^T \rangle = \frac{1}{2}
\begin{pmatrix}
    k(\tau) & -\mathcal{H}k(\tau) \\
    \mathcal{H} k(\tau) & k(\tau)
\end{pmatrix}.
\label{cross-covariance function, methods}
\end{equation}
We can obtain the \textit{cross-spectral density} $\tilde{\Theta}(\xi)$ of $\Psi(t)$ 
by taking the Fourier transform of $\Theta(\tau)$ using Eq.~\ref{hilbert transform fourier, methods}
\begin{equation}
\tilde{\Theta}(\xi) = \frac{1}{2} \tilde{k}(\xi)
\begin{pmatrix}
    1 & i~\textrm{sgn}(\xi)  \\
    -i~\textrm{sgn}(\xi)  & 1
\end{pmatrix}.
\label{cross-spectral density, methods}
\end{equation}
The conditional expectation $\langle \tilde{\beta}(\xi) \rangle_{\tilde{\alpha}(\xi)}$ can be obtained from the cross-spectral density as follows \citep{picinbono1996second}:
\begin{equation}
\langle \tilde{\beta}(\xi) \rangle_{\tilde{\alpha}(\xi)} = \gamma(\xi) \sqrt{\frac{\tilde{\Theta}_{22}(\xi)}{\tilde{\Theta}_{11}(\xi)}} \tilde{\alpha}(\xi) 
\label{fourier conditional expectation, methods}
\end{equation}
where the \textit{coherency} $\gamma(\xi)$ is  given by
\begin{equation}
\gamma(\xi) = \frac{\tilde{\Theta}_{12}(\xi)}{\sqrt{\tilde{\Theta}_{11}(\xi) \tilde{\Theta}_{22}(\xi)}} 
\label{coherency, method}
\end{equation}
Therefore, the real and the imaginary part of the signal are quadrature related in expectation if and only if the phase of the coherency $\arg[\gamma(\xi)]$ is equal to $-\textrm{sgn}(\xi) \frac{\pi}{2}$. For a cross-spectral density of the form given by Eq.~\ref{cross-spectral density, methods}, the coherency is
\begin{equation}
\gamma(\xi) = -i~\textrm{sgn}(\xi) = \exp{\big(-i ~\textrm{sgn}(\xi) \frac{\pi}{2}\big)}~.
\label{quadrature coherency, method}
\end{equation}
Using this result and the fact that $\tilde{\Theta}_{11}(\xi)$ is equal to $\tilde{\Theta}_{22}(\xi)$, it follows from Eq.~\ref{fourier conditional expectation, methods} that 
\begin{equation}
\langle \tilde{\beta}(\xi) \rangle_{\tilde{\alpha}(\xi)} = \exp{\big(-i ~\textrm{sgn}(\xi)\big)} \tilde{\alpha}(\xi)~,
\label{fourier quadrature conditional expectation, methods}
\end{equation}
Hence, the Fourier coefficients of the real and the imaginary part are quadrature related in expectation with $c(\xi) = |\gamma(\xi)| = 1$, where $c(\xi)$ is the scalar in Eq.~\ref{quadrature relation in expectation, methods}.

So far, we have shown that the CGPR with a quadrature covariance function allows to construct a probabilistic version of the analytic representation. However, this result does not fully realize our aim since the samples from the stochastic process are analytic functions with probability one. In order to understand this behavior, we will now study the eigenvalues and eigenvectors of the cross-spectral density. We denote the two eigenvectors of $\tilde{\Theta}(\xi)$ as $\boldsymbol{v}^{(1)}(\xi)$ and $\boldsymbol{v}^{(2)}(\xi)$ with eigenvalues $\lambda^{(1)}(\xi)$ and $\lambda^{(2)}(\xi)$ respectively. Note that, since the matrix is Hermitian and positive semi-definite, the eigenvalues are real and non-negative. For quadrature covariance functions, the eigenvectors are $\boldsymbol{v}^{(1)} = \frac{1}{\sqrt{2}}(i,1)$ and $\boldsymbol{v}^{(2)} = \frac{1}{\sqrt{2}}(-i,1)$ (for all values of $\xi$), and their corresponding eigenvalues are, respectively, $\lambda^{(1)}(\xi) = h(\xi) \tilde{k}(\xi)$ and $\lambda^{(2)}(\xi) = h(-\xi) \tilde{k}(\xi)$, where $h(\xi)$ is the Heaviside step function. Since the eigenvalues are never simultaneously non-zero, except for the irrelevant single point $\xi = 0$, all the bivariate complex distributions of $\tilde{\alpha}(\xi)$ and $\tilde{\beta}(\xi)$ are degenerate and the quadrature relationship is deterministic. Consequently, the resulting complex-valued signal does not alleviate the problems of the analytic representation in dealing with signals with overlapping amplitude and phase spectra.

\subsubsection{The periodic and the harmonic covariance functions}
\begin{figure}[!ht]
	\centering
    	\includegraphics[width=1.\textwidth] {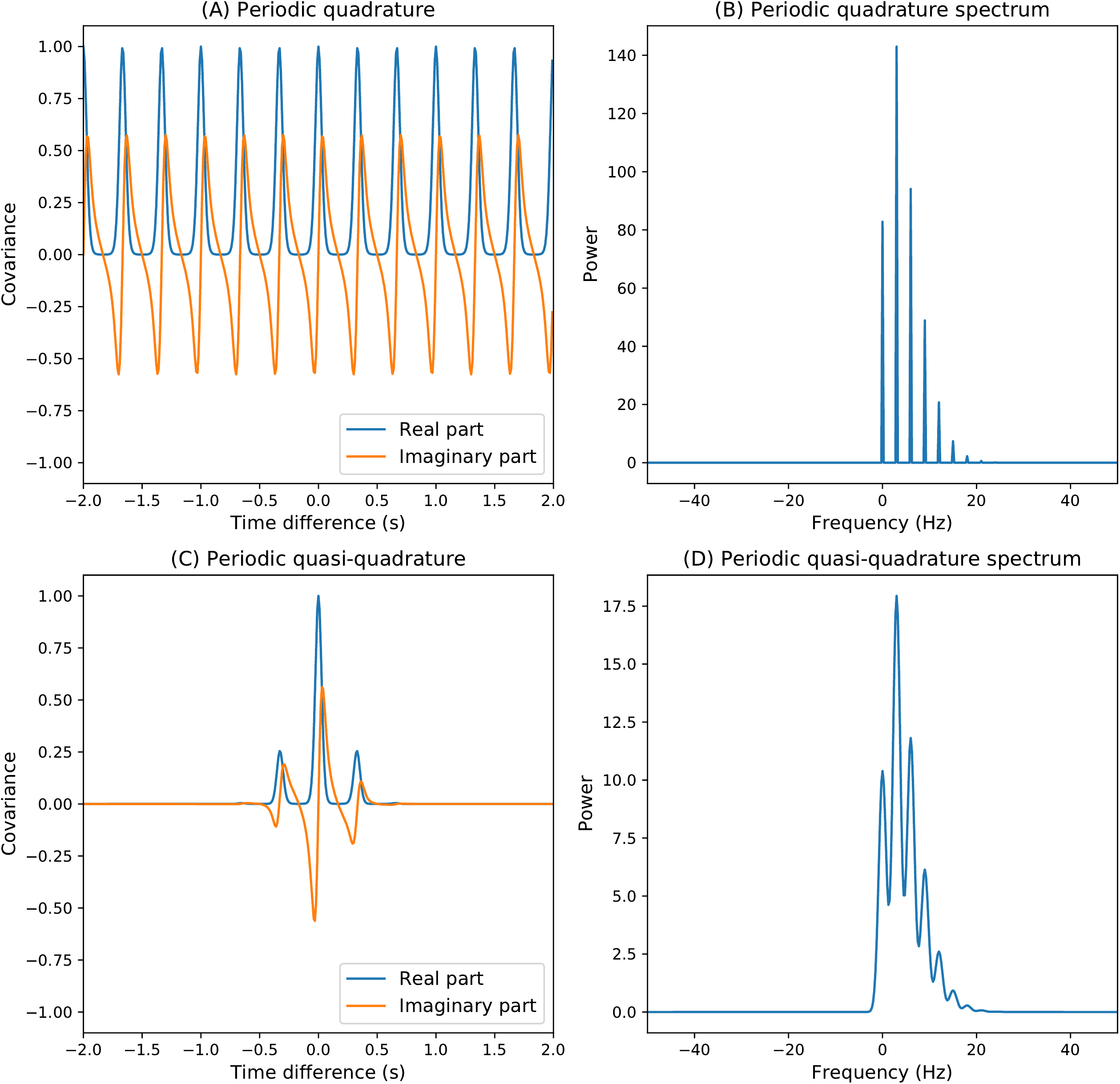}
	\caption{Quadrature and quasi-quadrature covariance functions. A) Real part (blue line) and imaginary part (green line) of the periodic quadrature covariance function ($\omega_0 = 2\pi \times 3\textrm{Hz}$, $\rho = 0.5$). The imaginary part was obtained by numerical Hilbert transform. B) Spectral density of the periodic quadrature covariance function. C) Real part (blue line) and imaginary part (green line) of the periodic quasi-quadrature covariance function ($\omega_0 = 2\pi \times 3\textrm{Hz}$, $\rho = 0.5$) with squared exponential envelope ($\mu = 0.2\textrm{s}$). D) Spectral density of the periodic quasi-quadrature covariance function with squared exponential envelope.}
	\label{figure 3}
\end{figure}

We will conclude this subsection by introducing two quadrature covariance functions that will be used in the remaining of the paper. One of them, the \emph{periodic covariance function}, is most suitable for the analysis of signals with sinusoidal waveforms. The other, the \emph{harmonic covariance function}, should be preferred in the analysis of non-sinusoidal signals. For a real-valued signal, the periodic covariance function is given by the following formula \citep{rasmussen2006gaussian}
\begin{equation}
k_p(\tau) = \exp{\big(-2 \sin^2(\frac{\omega_0}{2} \tau) / \rho^2 \big)},
\label{periodic covariance, methods}
\end{equation}
where $\omega_0$ is the center frequency and $\rho$ regulates the smoothness of the waveform. The associated quadrature covariance function can be obtained using Eq.~\ref{quadrature covariance function, methods}. Fig.~\ref{figure 3}A shows the real and the imaginary part of the periodic quadrature covariance function ($\omega_0 = 2\pi \times 3\textrm{Hz}$, $\rho = 0.5$). The imaginary part has been obtained from a numerical approximation to the Hilbert transform of the real part. Fig.~\ref{figure 3}B shows the spectral density of this covariance function. As expected, the spectral density is identically equal to zero for negative frequencies, implying that this covariance function is analytic. Since the function is periodic, the density concentrates on a sequence of discrete frequencies. Importantly, the spectral density has multiple peaks besides its main peak at ${\omega_0}/{2\pi}$ Hz. Consequently, when this covariance function is used in a GP regression on signals with non-sinusoidal waveforms, the waveforms of the resulting GP posterior expectations are non-sinusoidal themselves as the high-order harmonics are not suppressed. Unfortunately, the presence of high-order harmonics in the complex-valued signal compromises the estimation of the instantaneous phase. This happens because the quadrature filter is a linear operator and, when applied to a signal that is a linear superposition of sinusoidal waves, it outputs a signal that is the sum of the analytic representations of each harmonic component. Consequently, the resulting instantaneous phase is influenced by the phase of several high frequency components that determine the signal waveform but whose phase cannot be meaningfully interpreted as the instantaneous phase of the signal. 

In order to find a covariance function that suppresses higher order harmonics, it is useful to analyze the asymptotic behavior of the periodic covariance function for the smoothness parameter $\rho^2$ tending to infinity (which corresponds to increasing smoothness). Specifically, the following can be shown: 
\begin{equation}
\exp{\big(-2 \sin^2(\frac{\omega_0}{2} \tau) / \rho^2 \big)} \sim 1 -2 \sin^2(\frac{\omega_0}{2} \tau ) / \rho^2 = (1 - 1/\rho^2) + \cos(\omega_0 \tau)/\rho^2~
\label{asymptotic relation, results}
\end{equation}
The symbol $\sim$ in the relation $f(x) \sim g(x)$ denotes that the limit of the ratio $f(x)/g(x)$ (for $x$ approaching some specified value) equals 1. The result in Eq. \ref{asymptotic relation, results} follows from the following two facts: (1) the elementary asymptotic relation $\exp(c/\rho^2) \sim 1 + c/\rho^2$ for $\rho^2 \rightarrow \infty$ (where $c$ is an arbitrary constant), and (2) the trigonometric identity $\sin^2(x) = (1 - \cos(2x))/2$. The leading term of this expansion is the constant covariance function $(1 - 1/\rho^2)$ that models a constant shift in the mean of the signal. More interestingly, the first order term $\cos(\omega_0 \tau)/\rho^2$ is a pure sinusoidal component with a spectral peak at ${\omega_0}/{2\pi}$. By applying the quadrature filter to to this expression, we obtain the first order asymptotic expansion of the quadrature periodic covariance function:
$$\mathcal{A} k_p(\tau) \sim (1 - 1/\rho^2) + \exp{\big(i \omega_0 \tau \big)}/\rho^2~.$$
In obtaining this result, we made use of the fact that the Hilbert transform of a constant is always zero and, therefore the constant term is not affected by the quadrature filter. As a result, up to a constant and a scaling term, the quadrature periodic covariance function is asymptotically related to the following harmonic covariance function:
\begin{equation}
	\mathcal{A} k_h(\tau) = \mathcal{A} \cos(\omega_0 \tau) = \exp{\big(i \omega_0 \tau\big)}~.
\label{harmonic covariance, results}
\end{equation}
This covariance function can be constructed from the outer product of a single basis function: $\Phi(t) = \exp(i \omega_0 t)$. In fact,
$$\mathcal{A} k_h(\tau) = \Phi(t_1) \conj{\Phi(t_2)}~, $$
where $\tau = t_2 - t_1$. Consequently, performing a GP regression using this covariance function is equivalent to a linear fit of a single complex sinusoid \citep{rasmussen2006gaussian}. 

\subsection{Quasi-quadrature covariance functions}
We will now define a family of non-degenerate covariance functions that generate processes whose real and imaginary parts have quadrature relationship in expectation. To this end, we have to modify quadrature covariance functions such that both eigenvalues of the cross-spectral density differ from zero for all frequencies while the phase of the coherency remains identically equal to $-\textrm{sgn}(\xi) \frac{\pi}{2}$ (see Eq.~\ref{fourier conditional expectation, methods}). In addition, we need to show that, as the frequency $\xi$ tends to infinity, the quadrature relationship becomes deterministic, meaning that $c(\xi)$ tends to one and that the variance of ${\langle \tilde{\beta}(\xi) \rangle}_{\tilde{\alpha}(\xi)}$ tends to zero. The motivation for this requirement is that we aim to recover the behavior of the quadrature filter for the frequency range over which the overlap between the clockwise and the counterclockwise parts is usually negligible, which is for the high frequencies (see Fig.~\ref{figure 2}). 

\subsubsection{Definition and general results}
We begin by defining quasi-quadrature covariance functions as the product of a long time-scale envelope function $f(\tau)$ and a quadrature covariance function. More formally, we define the quasi quadrature operator $\mathcal{A}_f(\tau)$ as follows:
\begin{equation}
\mathcal{A}_f k(\tau) = f(\tau) \mathcal{A} k(\tau)~.
\label{quasi-quadtrature covariance, method}
\end{equation}
The envelope function has to be even and positive definite, meaning that its Fourier transform is real and positive-valued, so that the result is a valid covariance function. Moreover, because we want to recover the behavior of the quadrature filter for high frequencies, the Fourier transform of the envelope function must have the following properties:
\begin{equation}
\lim_{\xi\to\infty} \tilde{f}(\xi) = 0 \\
\label{envelope requirements I, method}
\end{equation}
and
\begin{equation}
\forall \epsilon > 0,~~ \lim_{\xi\to\infty} \frac{\tilde{f}(\xi - \epsilon)}{\tilde{f}(\xi)} = \infty.
\label{envelope requirements II, method}
\end{equation}
We will now show that the eigenvalues of quasi quadrature covariance functions are simultaneously different from zero. To this end, we compute the cross-spectral density $\tilde{\Theta}_f(\xi)$ of the new process by applying the Fourier convolution theorem on its cross-covariance function $\Theta_f(\tau) = f(\tau) \Theta(\tau)$. The result is
\begin{equation}
\tilde{\Theta}_f(\xi) = \tilde{f}(\xi)* \tilde{\Theta}(\xi) = \int_{-\infty}^{+\infty} \tilde{f}(\xi - \upsilon)~ \tilde{k}(\upsilon)
\begin{pmatrix}
    1 & -i~\textrm{sgn}(\upsilon)  \\
    i~\textrm{sgn}(\upsilon)  & 1
\end{pmatrix}
d\upsilon.
\label{cross-spectral density quasi-quadrature, methods}
\end{equation}
This is an integral of matrices that share the same eigenvectors. Consequently, the eigenvectors of $\tilde{\Theta}_f(\xi)$ are still $\boldsymbol{v}^{(1)}$ and $\boldsymbol{v}^{(2)}$ for all frequencies but the eigenvalues are now given by
\begin{equation}
{\lambda_f}^{(1,2)}(\xi) = \int_{-\infty}^{+\infty} \tilde{f}(\xi - \upsilon) \lambda^{(1,2)} (\upsilon) dv = \int_{-\infty}^{+\infty} \tilde{f}(\xi - \upsilon) \tilde{k}(\upsilon) h(\pm \upsilon) d\upsilon.
\label{eigenvalue quasi-quadrature, methods}
\end{equation}
Therefore, the first and the second eigenvalue functions are given by the convolution of, respectively, the analytic ($\tilde{k}(\xi) h(+ \xi)$) and the anti-analytic ($\tilde{k}(\xi) h(- \xi)$) part of the spectrum $\tilde{k}(\xi)$ with the Fourier transform of $f(\tau)$. Note that, since $f(\tau)$ is positive definite, $\tilde{f}(\xi)$ is strictly positive and both eigenvalues are never exactly zero. Consequently, the cross-spectral density is never degenerate, and the Fourier coefficients of the real and the imaginary part are not perfectly correlated.

We still need to show that quasi-quadrature covariance functions define complex-valued GPs whose real part and imaginary part have a quadrature relationship in expectation. It follows from Eq.~\ref{quadrature relation in expectation, methods} and Eq.~\ref{fourier conditional expectation, methods} that this requirement (quadrature relationship in expectation) implies that the argument of the coherency has to be equal to $-i~ \textrm{sgn}(\xi)$. By direct calculation, we can show that the cross-spectrum (the numerator of coherency) can be expressed as a linear combination of the eigenvalues (see appendix II):
\begin{equation}
\Theta(\xi)_{12} = a {\lambda}^{(1)}(\xi) + b {\lambda}^{(2)}(\xi)~,
\label{coherency from eigendecomposition}
\end{equation}
Crucially, the complex numbers $a$ and $b$ solely depend on the eigenvectors $\boldsymbol{v^{(1)}}$ and $\boldsymbol{v^{(2)}}$. From this expression, together with the fact that the denominator of coherency is a positive real number, it follows that $\arg[\gamma(\xi)]$ does not change if the eigenvalues are multiplied by a positive constant while the eigenvectors are left unchanged. More generally, $\arg[\gamma(\xi)]$ is left unchanged when both the positive-valued eigenvalue functions ${\lambda}^{(1)}(\xi)$ and ${\lambda}^{(2)}(\xi)$ are convolved with the same positive-valued function. Therefore, Eqs. ~\ref{cross-spectral density quasi-quadrature, methods} and~\ref{eigenvalue quasi-quadrature, methods} imply that $\arg[\gamma(\xi)]$ is the same for quadrature and quasi-quadrature covariance functions. Consequently, the quadrature relationship holds in expectation. We can obtain a simple formula for the coherency $\gamma(\xi)$ from Eq.~\ref{coherency, appendix II} in appendix II:
\begin{equation}
\gamma(\xi) = -i \frac{\lambda_f^{(1)}(\xi) - \lambda_f^{(2)}(\xi)}{\lambda_f^{(1)}(\xi) + \lambda_f^{(2)}(\xi)},
\label{quasi-quadrature coherence}
\end{equation}
As expected, this formula reduces to Eq.~\ref{quadrature coherency, method} when the envelope function is constant and the resulting process is in exact quadrature. Furthermore, as proven in Appendix III, when the envelope function respects the requirement given in Eq.~\ref{envelope requirements II, method} we have that
$$
\lim_{\xi\to\infty} \frac{\lambda_f^{(2)}(\xi)}{\lambda_f^{(1)}(\xi)} = 0~.
$$
Consequently, from Eq.~\ref{quasi-quadrature coherence} we obtain
$$
\lim_{\xi\to\infty} \gamma(\xi) = -i~.
$$
Therefore, the Fourier coefficients of a stochastic process with quasi-quadrature covariance function are in almost exact quadrature relationship at high frequencies. 

We gain further insight into the deviation from exact quadrature relationship between the real and the imaginary parts of quasi-quadrature GPs by studying the variance of the conditional distribution $p(\tilde{\beta}(\xi)|\tilde{\alpha}(\xi))$ as a function of the frequency $\xi$. As explained in the previous subsection, this is a complex normal distribution. Its conditional expectation, as given in Eq.~\ref{fourier conditional expectation, methods}, played a crucial role in proving that the real and imaginary parts of quadrature GPs have a quadrature relationship. We will now show that, for large values of $\xi$, the random variable $\beta(\xi)|\alpha(\xi)$ is strongly concentrated around its expected value. We will denote the deviation of $\beta(\xi)|\alpha(\xi)$ from its expected value as $\eta(\xi)$. The complex variable $\eta(\xi)$ is circularly-symmetric as it has zero mean and zero relation matrix \citep{picinbono1996second}. Moreover, its variance is \citep{picinbono1996second}:
\begin{equation}
\chi(\xi) = \Theta_{11}(\xi) - \frac{|\Theta_{21}(\xi)|^2}{\Theta_{22}(\xi)} = 4 \frac{\lambda^{(1)}(\xi)\lambda^{(2)}(\xi)}{\lambda^{(1)}(\xi) + \lambda^{(2)}(\xi)} \sim 4 \lambda^{(2)}(\xi)~,
\label{conditional variance, method}
\end{equation}
where the asymptotic relation holds for $\omega \rightarrow +\infty$. We can quantify the deviation from exact quadrature relationship as the modulus of $\eta(\xi)$. The probability of $|\eta(\xi)|$ being smaller than an arbitrary positive number $\epsilon$ is:
$$
P\big(|\eta(\xi)|< \epsilon \big) = \frac{1}{\pi \chi(\xi)}\int_{C(\epsilon)} \exp{\big(-z\conj{z}/\chi(\xi)\big)} d\mathfrak{R}z~,
$$
where $C(\epsilon)$ denotes a circle centered at the origin of the complex plane with radius $\epsilon$. This integral can be solved by changing variables to polar coordinates, the result is:
$$
P\big(|\eta(\xi)|< \epsilon \big) = 1 - \exp{\big(-\epsilon^2/\chi(\xi)\big)}~,
$$
Consequently, the probability of the deviation $|\eta(\xi)|$ to be bigger than $\epsilon$ is 
\begin{equation}
P\big(|\eta(\xi)|> \epsilon \big) = 1 - P\big(|\eta(\xi)|< \epsilon \big) = \exp{\big(-\epsilon^2/\chi(\xi)\big)}~.
\label{probability bound, method}
\end{equation}
Crucially, $\chi(\xi)$ vanishes as $\xi \rightarrow \infty$ and, consequently
$$
\forall \epsilon >0, \lim_{\xi\to\infty} P\big(|\eta(\xi)|> \epsilon \big) = 0~,
$$
as had to be demonstrated.

\subsubsection{The harmonic quasi-quadrature covariance function}
The behavior of quasi-quadrature covariance functions can be more easily understood by considering the special case of the harmonic covariance function, as given in Eq.~\ref{harmonic covariance, results}, with squared exponential envelope:
\begin{equation}
f(\tau) = \exp{\big(-\frac{\tau^2}{2 \mu^2}\big)},
\label{squared exponential envelope, method}
\end{equation}
where $\mu$ is its characteristic time scale. This case is particularly simple since the spectrum of the harmonic covariance function is $$k(\xi) = \delta(\upsilon + \omega_0) + \delta(\upsilon - \omega_0)~.$$
From Eq.~\ref{eigenvalue quasi-quadrature, methods}, the eigenvalues of the cross-spectral density are:
$$
{\lambda_f}^{(1)}(\xi) = \mu \int_{-\infty}^{+\infty} \exp{\big(-\mu^2(\xi - \upsilon)/2\big)} \delta(\upsilon + \omega_0) d\upsilon = \mu \exp{\big(-\mu^2 (\xi - \omega_0)/2\big)}
$$
and
$$
{\lambda_f}^{(2)}(\xi) = \mu \int_{-\infty}^{+\infty} \exp{\big(-\mu^2(\xi - \upsilon)/2\big)} \delta(\upsilon - \omega_0) d\upsilon = \mu \exp{\big(-\mu^2 (\xi + \omega_0)/2\big)},
$$
since the Fourier transfoirm of the squared exponential envelope is $\mu \exp{(-\mu^2 \xi^2)}$. Plugging this expression in Eq.~\ref{squared exponential envelope, method}, we obtain a very simple formula for the coherency:
\begin{equation}
\gamma(\xi) = -i \frac{\exp{\big(-\mu^2 (\xi - \omega_0)/2\big)} - \exp{\big(-\mu^2 (\xi + \omega_0)/2\big)}}{\exp{\big(-\mu^2 (\xi - \omega_0)/2\big)} + \exp{\big(-\mu^2 (\xi + \omega_0)/2}\big)} = -i \tanh{\big( 2 \mu^2 \omega_0 ~ \xi \big)}.
\label{harmonic_gaussian coherency, method}
\end{equation}
This formula is strikingly similar to the formula for the coherency of quadrature covariance functions in Eq.~\ref{quadrature coherency, method}. In fact, the formula can be obtained from Eq.~\ref{quadrature coherency, method} by replacing the discontinuous Heaviside function by the smooth hyperbolic tangent. 

\subsubsection{Visualizations}
\begin{figure}[!ht]
	\centering
    	\includegraphics[width=1.\textwidth] {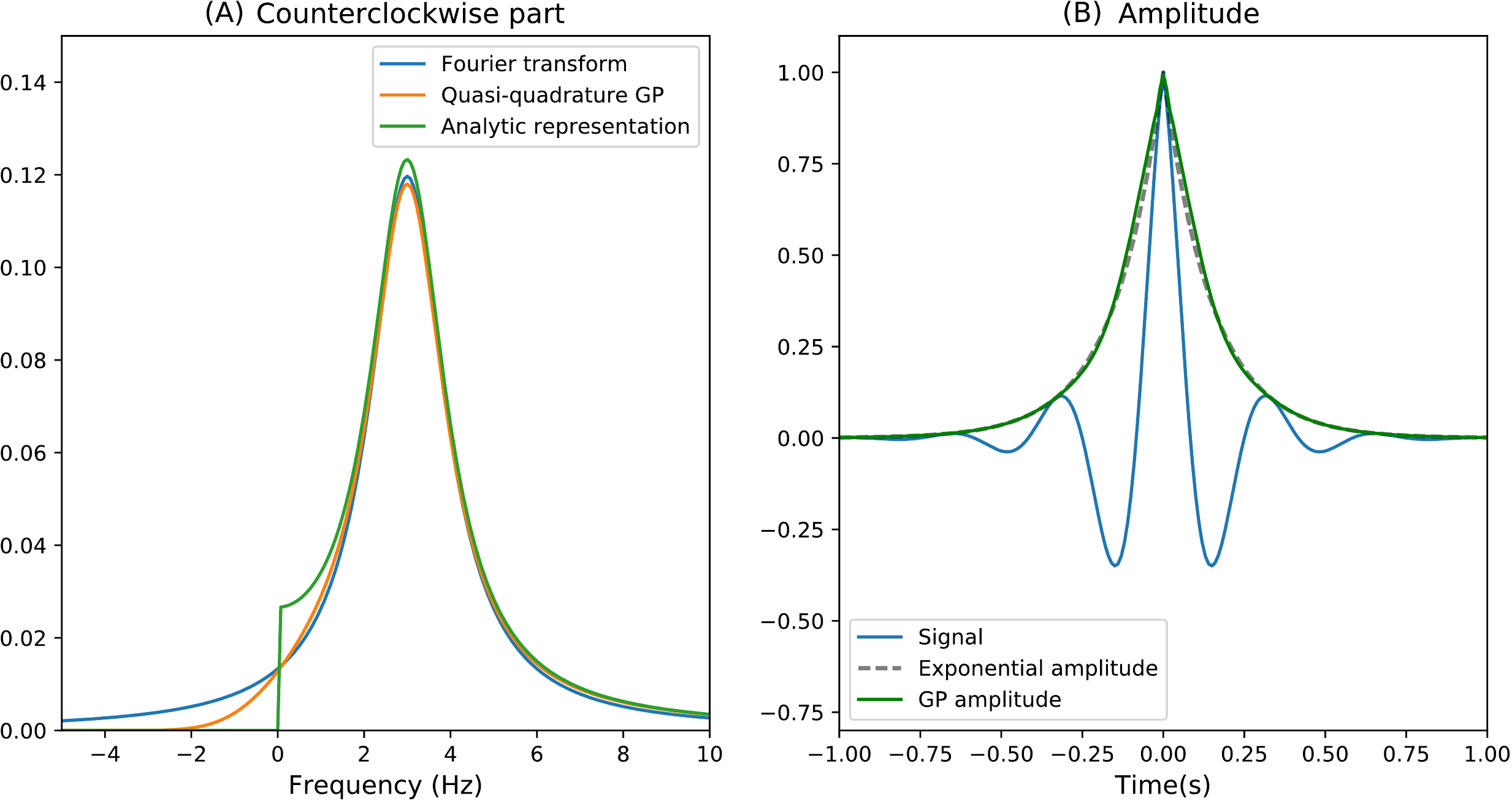}
	\caption{ Quasi-quadrature CGPR analysis of the example signal. A) Clockwise part (blue), analytic representation (green) and quasi-quadrature CGPR signal (orange) in the frequency domain. B) Signal envelope (dashed), quasi-quadrature CGPR envelope (green).}
	\label{figure Extra}
\end{figure}

To improve our intuition about quasi-quadrature covariance functions, we now visualize one of the covariance functions used in this paper. Fig.~\ref{figure 3}C shows the quasi-quadrature covariance function that is obtained by multiplying a quadrature periodic covariance function with a squared exponential envelope ($\mu = 0.2\textrm{s}$). The envelope breaks the exact periodicity of the function, thereby allowing to model signals that are not exactly periodic. Fig.~\ref{figure 3}D shows the spectrum of this covariance function. As a result of the multiplication by the envelope function, the negative frequencies are not exactly zero. Fig.~\ref{figure Extra} shows the results of a quasi-quadrature CGPR analysis (periodic, Gaussian envelope, $\omega_0 = \pi $, $\rho = 0.1$, $\mu = 0.2$) of the example signal given in Eq.\ref{test signal, methods}. Panel A shows the absolute value of the resulting complex-valued signal in the frequency domain. Note that this signal has non-vanishing energy at negative frequencies and fit the clockwise part more closely than the analytic representation. Panel B shows the real and the imaginary part of the signal, together with the constructed instantaneous amplitude. 

\section{Connections between quasi-quadrature CGPR and Morlet wavelet analysis}
We can now show that Morlet wavelet analysis is a special case of quasi-quadrature CGPR in the high noise limit. Wavelet analysis is usually applied on discrete-time data by discrete the convolution operator in Eq.~\ref{wavelet convolution, methods}. The formula for the discretized wavelet analysis is
\begin{equation}
\boldsymbol{s}_w = W \boldsymbol{s}~,
\label{discrete wavelet, method}
\end{equation}
where each row of the matrix $W$ is a translated version of the original wavelet. In particular, the elements of $W$ are given by
\begin{equation}
W_{jk} = \exp{\big(-\frac{(t_k - t_j)^2}{2 \mu^2}\big)} ~ \exp{\big(i \omega_0 (t_k - t_j)\big)}~.
\label{wavelet matrix, method}
\end{equation}
This matrix is identical to the covariance matrix obtained from the harmonic quasi-quadrature covariance function with a squared exponential envelope (see Eq.~\ref{harmonic covariance, results} and Eq.~\ref{squared exponential envelope, method}). Using Eq.~\ref{complex posterior expectation GP, methods} and assuming a white noise observation model, we obtain that the complex-valued signal obtained using the associated CGPR is given by the following formula:
\begin{equation}
\boldsymbol{m}_{\zeta|s} = W(\mathfrak{R} W + \lambda I)^{-1}  \boldsymbol{s}~.
\label{quasi-quadrature posterior expectation, method}
\end{equation}
The connection between quasi-quadrature CGPR and Morlet wavelet analysis is now apparent. In the limit $\lambda \rightarrow \infty$, we obtain the following (element-wise) asymptotic relation:
\begin{equation}
W(\mathfrak{R} W + \lambda I)^{-1}  \boldsymbol{s} \sim \lambda^{-1}  W \boldsymbol{s} \propto \boldsymbol{s}_w~.
\label{quasi-quadrature wavelet equivalence, method}
\end{equation}
Therefore, Morlet wavelet analysis differs from the high noise limit of a quasi-quadrature CGPR analysis only by a scale factor, which does not contribute to the instantaneous phase. Hence:
\begin{equation}
\arg[{\boldsymbol{s}_w}] = \lim_{\lambda \to\infty} \arg[\boldsymbol{m}_{\zeta|s}]~. 
\label{wavelet_phase_equivalence, method}
\end{equation}
From this analysis, we can conclude that the quasi-quadrature CGPR approach extends the Morlet wavelet analysis to low and intermediate noise settings.

\section{Experiments}
We begin this section by comparing the performance of CGPR, analytic representation and Morlet wavelet analysis in estimating instantaneous amplitude and phase of different kinds of stochastic oscillations. Because the quadrature filter (required for the construction of the analytic representation) is destabilized by broadband noise, these simulations have been performed using noise-free data. Subsequently, we compare the performance of CGPR and Morlet wavelet analysis using noise-corrupted data. Finally, we report on the analysis of real magneto-encephalograpic (MEG) data. 

\subsection{Analysis of deterministic chirplets}
A chirplet is a signal whose instantaneous frequency increases as a function of time and whose amplitude is concentrated in a small time window\citep{yang2013multicomponent}. The analysis of chirplets is relevant in many scientific fields. For example, gravitational waves have a chirplet waveform \citep{adams2013gravitational}. 

We generate chirplets by multiplying a cosine wave (with either linear, quadratic or exponential frequency increase) with an amplitude envelope. 
\begin{equation} 
s(t) = h(t) t^k \exp{\big(- \frac {t} {b}\big)} \cos \Phi(t)
\label{chirplet envelope, results}
\end{equation}
where $k$ is an integer from 1 to 3, $b$ is a parameters that regulates the width of the envelope and $h(t)$ is the Heaviside step function. The instantaneous phase $\Phi(t)$ is given by $(\omega_0 + a t) t$ (linear frequency growth), $(\omega_0 + a t^2) t$ (quadratic frequency growth) or $(\omega_0 + a \exp{(t/2)}) t$ (exponential frequency growth) plus a random initial phase. In total, we have nine possible combinations of envelope and phase functions, each having the width $b$, the initial frequency $\omega_0$ and the frequency factor $a$ as parameters. We run a simulation study by randomly selecting 500 of these combinations, with the parameters sampled from the uniform distributions $b \in [0.1 ,~0.3]$, $2 \pi \omega_0 \in [0.1,~2]$, $k \in \{1,2,3\}$ and $2 \pi a \in [0.1,~0.4]$. For each of these chirplets, we extracted instantaneous amplitude and frequency using both the analytic representation and the GP-based complex signal. The latter was obtained using a periodic quasi-quadrature covariance with squared exponential envelope and parameters $\omega_0 = 0.5 \text{Hz}$, $\rho = 0.1$ and $\mu = 0.3 \text{s}$. In this analysis, the signal did not contain any noise, and therefore we set the noise covariance matrix $I \lambda$ in Eq.~\ref{complex posterior expectation GP, methods} to be close to zero ($\lambda = 10^-6$). For the Morlet wavelet analysis, we used $\omega_0 = 2 \pi \times 2$ and $\mu = 0.2$.

\begin{figure}[!ht]
	\centering
    	\includegraphics[width=1.\textwidth] {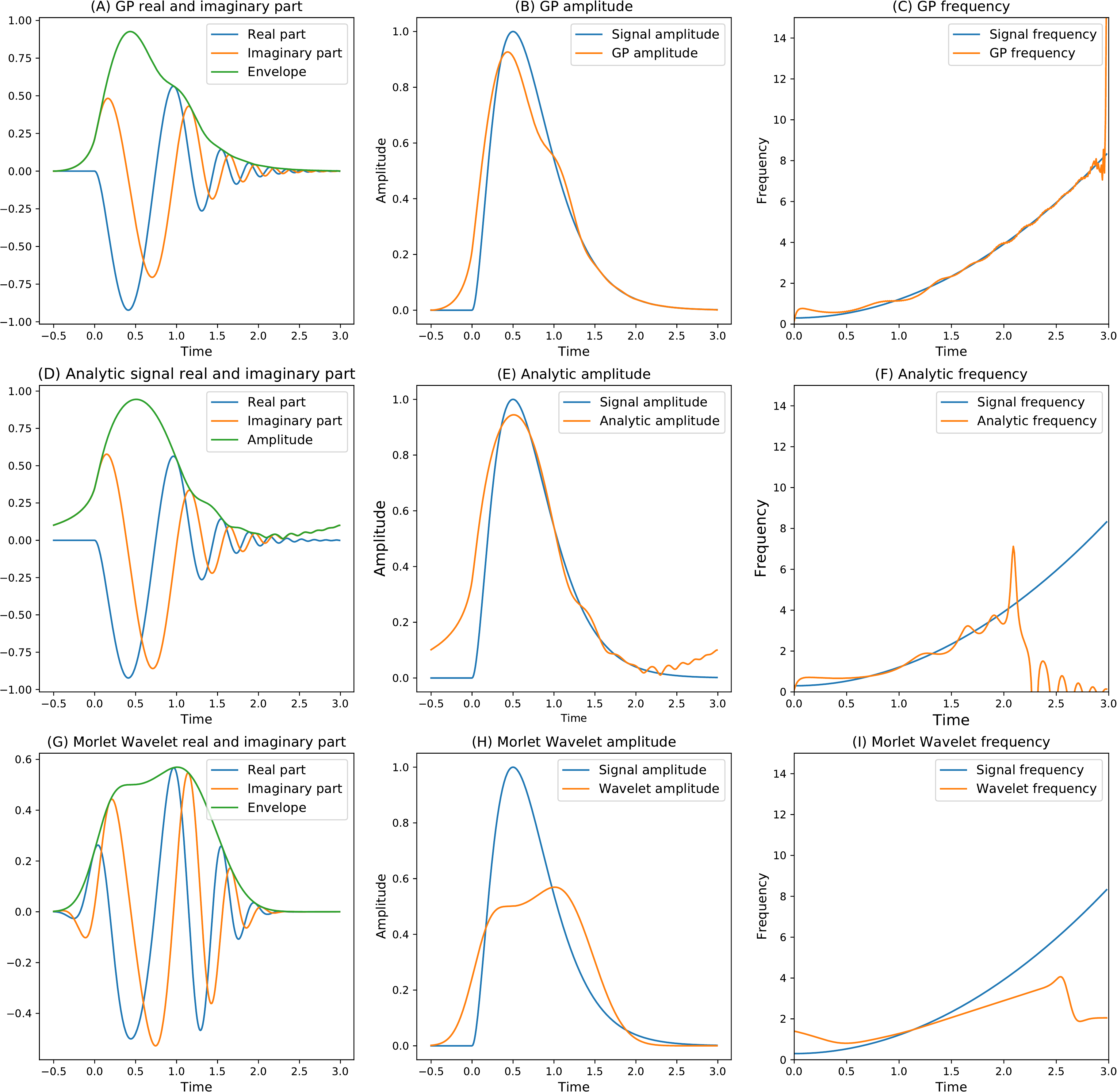}
	\caption{ Analysis of a chirplet. A) Real part, imaginary part and amplitude of the CGPR complex signal. B,C) Instantaneous amplitude and frequency obtained through CGPR analysis. D) Real part, imaginary part and amplitude of the analytic representation. E,F) Instantaneous amplitude and frequency obtained through the analytic representation. G) Real part, imaginary part and amplitude of the wavelet signal. H,I) Instantaneous amplitude and frequency obtained using wavelet analysis.}
	\label{figure 4}
\end{figure}

Fig.~\ref{figure 4} shows the results for a representative chirplet. Panel A shows the real and imaginary parts of the CGPR-based complex signal together with its amplitude. Note that, since the CGPR was noise free, the real part is identical to the original signal. Panel B and C compare, respectively, the amplitude and frequency obtained through CGPR with the "ground truth" given by $\mathfrak{A}(t) = t^k \exp{(- \frac {t} {b})}$ and $\mathfrak{P}(t) = \Phi(t)$ (see Eq.~\ref{chirplet envelope, results}). The CGPR method accurately recovers both instantaneous amplitude and frequency, except for a large deviation of the estimated frequency at the very end of the chirplet. This is probably due to the fact that the chirplets are constructed such that the amplitude of the signal becomes very small at the end of the trial. As a consequence, it becomes almost impossible to estimate the frequency. Panels D, E and F show the same results for the analytic representation. We can see that the analysis with the analytic representation fails to estimate the frequency in the low-amplitude/high-frequency segment and this failure is accompanied by ringing of the estimated amplitude envelope. In addition, the analytic method exhibits serious boundary effects even though the signal is very close to zero at the boundaries. Finally, panels G, H and I show the results of the Morlet wavelet analysis. In this case, the real part of the resulting complex-valued signal is not identical to the original signal. The band-pass filter that is intrinsic to the wavelet analysis produces visible distortions in the estimated envelope. The suppression of the very high frequency components of the chirlplet also leads to underestimation of the instantaneous frequency in the right side of the plot. 

\begin{figure}[!ht]
	\centering
    	\includegraphics[width=1.\textwidth] {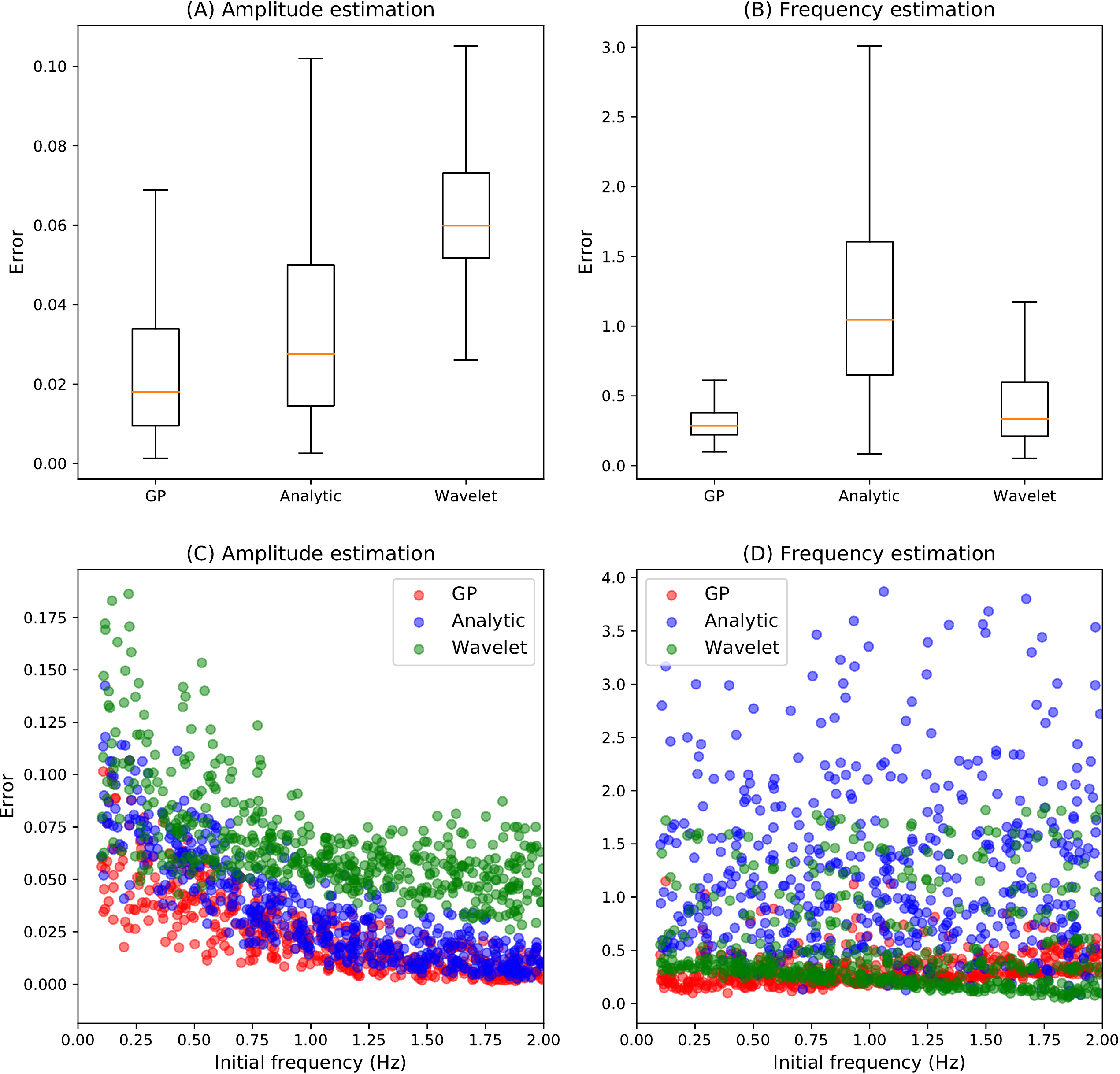}
	\caption{Comparative analysis of deterministic chirplets. (A,B) Comparison between GP-based complex signal, analytic representation and Morlet wavelet analysis. (C,D) Performance of the methods as function of the (random) initial frequency $\frac{d \mathfrak{P}}{dt}\Bigr|_{t = 0}$ of the randomly generated ground truth chirplets.}
	\label{figure 5}
\end{figure}

Fig.~\ref{figure 5} shows the outcome of the comparison between the methods. Errors were quantified as mean absolute deviation from the ground truth. The boxplots in panels A and B show the comparison between GP-based complex signal, analytic representation and Morlet wavelet analysis for, respectively, amplitude and frequency. We tested the differences between the methods using pairwise Wilcoxin signed rank tests. 

CGPR outperforms the alternative methods in both amplitude and frequency estimation (p \textless~0.01). The instantaneous amplitude functions obtained using the analytic representation are more reliable than those obtained using wavelet analysis (p \textless~0.01). However, the Morlet wavelet method produces more accurate instantaneous frequency functions than the analytic representation (p \textless~0.01). Panels C and D show the performance of the methods as a function of the initial frequency. For amplitude estimation, the difference between the methods is the largest for very low frequencies. 

\subsection{Analysis of stochastic oscillatory processes}
We now compare the methods with respect to their ability to reconstruct amplitude and frequency of a stochastic oscillatory signal. We generate stochastic oscillations by multiplying a smooth random amplitude function $\mathfrak{A}(t)$ with a sinusoidal wave with a smooth random frequency function $\mathfrak{P}(t)$. Specifically, the oscillatory signal $s(t)$ was generated as  follows:
\begin{equation}
s(t) = \mathfrak{A}(t) \cos(\mathfrak{P}(t) + \phi_0)~.
\label{simulated stochastic signal, results}
\end{equation}
In this formula, the amplitude process $\mathfrak{A}(t)$ is given by a positive-valued non-linear transform of a GP:
\begin{equation}
\mathfrak{A}(t) = \sqrt{x(t)^2 + 1} - 1,
\label{amplitude simulated stochastic signal, results}
\end{equation}
in which $x(t)$ is a GP with a mean 
identically equal to $\sqrt{3}$ and a covariance that is a squared exponential with time-scale $\mu_{x}$: 
$$k_{x}(\tau) = \exp{\big(- \frac{\tau^2}{2 {\mu_{x}^2}}\big)}~.$$
The instantaneous phase function $\mathfrak{P}(t)$ is a GP with mean $\omega_0 t$ (the phase of a stationary oscillation with frequency $\omega_0$) and squared exponential covariance function with time-scale $\mu_{\mathfrak{P}}$: 
$$k_{\mathfrak{P}}(\tau) = \exp{\big(- \frac{\tau^2}{2 {\mu_{\mathfrak{P}}^2}}\big)}~.$$
Finally, the initial phase $\phi_0$ is a uniform random variable ranging from $-\pi$ to $\pi$. Furthermore, the samples of the oscillatory process were multiplied by a Hann taper in order to reduce the boundary artifacts of the analytic representation. The instantaneous frequency of the stochastic oscillation was defined as the first temporal derivative of the instantaneous phase, i.e. $\frac{d \mathfrak{P}}{dt}$.

We ran a simulation study by generating 500 stochastic oscillatory signals. Each signal was generated by randomly initializing the parameters $\mu_{x}$, $\omega_0$ and $\mu_{\mathfrak{P}}$, sampling $x(t)$ and $\mathfrak{P}(t)$ from the respective multivariate Gaussian distributions and combining them using Eq.~\ref{simulated stochastic signal, results} and \ref{amplitude simulated stochastic signal, results}. The ranges of the parameters were $\mu_x \in [0.1 ,~0.4]$, $2 \pi \omega_0 \in [1,~3]$ and $\mu_{\mathfrak{P}} \in [0.1 ,~0.4]$. 

We analyzed the resulting time series using quasi-quadrature periodic CGPR ($\omega_0 = 2 \pi \times 0.5$, $\rho = 0.1$, $\mu = 0.3$, $\lambda = 10^{-4}$), analytic representation and Morlet wavelet analysis ($\omega_0 = 2 \pi \times 2$, $\mu = 0.2$). 

\begin{figure}[!ht]
	\centering
    	\includegraphics[width=1.\textwidth] {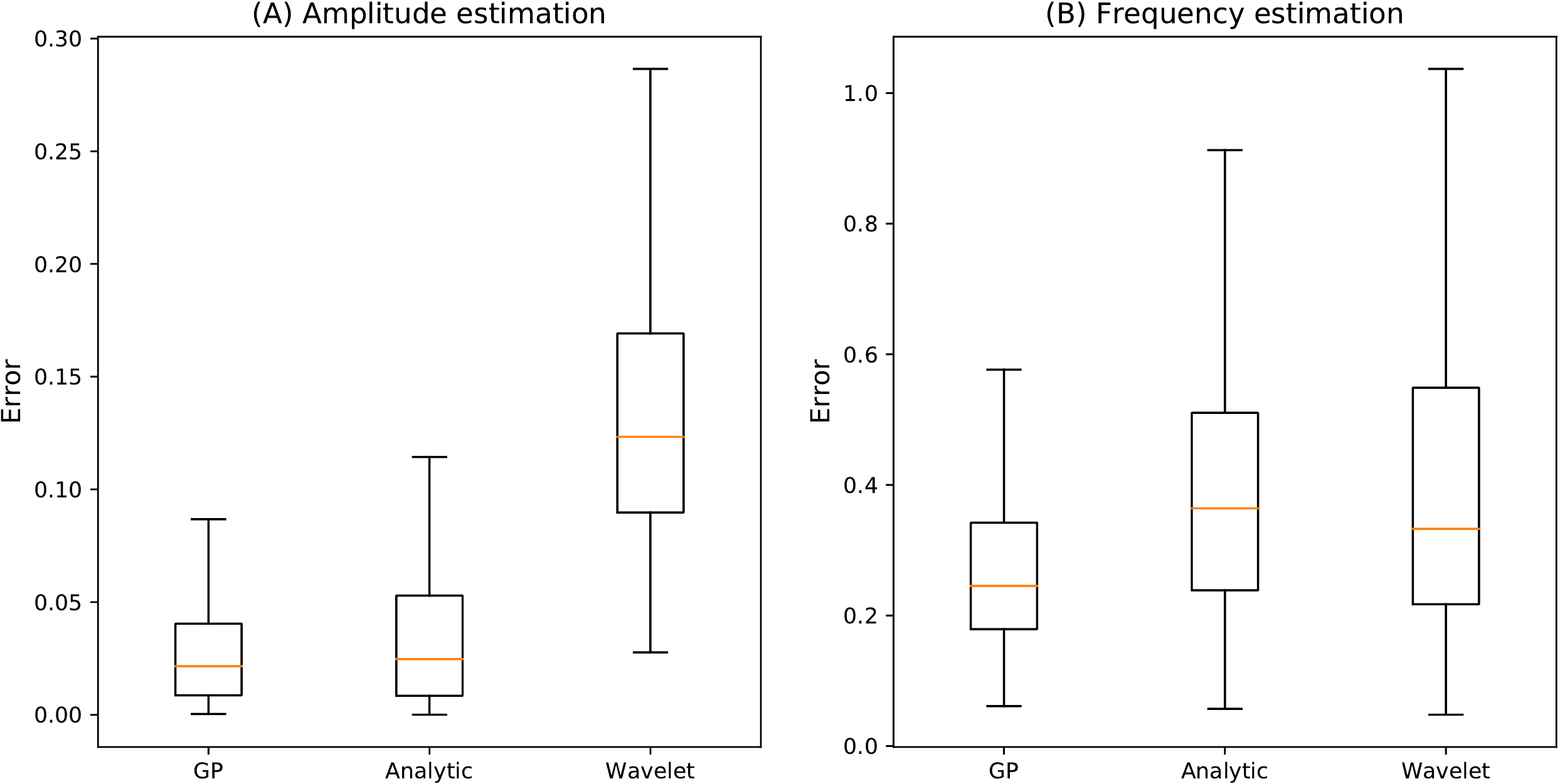}
	\caption{Comparative analysis of stochastic oscillations. (A,B) Comparison between GP-based complex signal, analytic representation and Morlet wavelet analysis.}
	\label{figure 6}
\end{figure}

Fig.~\ref{figure 6} shows the results of the simulation study. The complex GP signal outperforms the other methods (p \textless~0.01), although the difference with the analytic representation is less pronounced than in the previous simulation. Again, the Morlet wavelet method recovers instantaneous amplitude functions that are less reliable than those obtained using the analytic representation (p \textless~0.01). However, it recovers instantaneous frequency functions that are more reliable than those obtained using the analytic representation (p \textless~0.01). 

\subsection{Analysis of noise corrupted signals}
We now validate the robustness of our CGPR method on noise-corrupted data. We compare the robustness of two different covariance functions: periodic and harmonic. Furthermore, we compare the results with the Morlet wavelet analysis. We do not include a comparison with the analytic representation because the quadrature filter is easily destabilized by broad-band noise. We generated synthetic stochastic oscillations using the method explained in the previous subsection. These signals were then corrupted by Gaussian-white noise (with mean 0 and standard deviation 0.1). We ran a simulation study by generating 500 noise-corrupted signals. The resulting time series were analyzed using quasi-quadrature harmonic CGPR ($\omega_0 = 2 \pi \times 2$, $\mu = 0.2$, $\lambda = 0.8$), quasi-quadrature periodic CGPR ($\omega_0 = 2 \pi \times 0.5$, $\rho = 0.1$, $\mu = 0.3$, $\lambda = 0.8$) and Morlet wavelet analysis ($\omega_0 = 2 \pi \times 2$, $\mu = 0.2$).

\begin{figure}[!ht]
	\centering
    	\includegraphics[width=1.\textwidth] {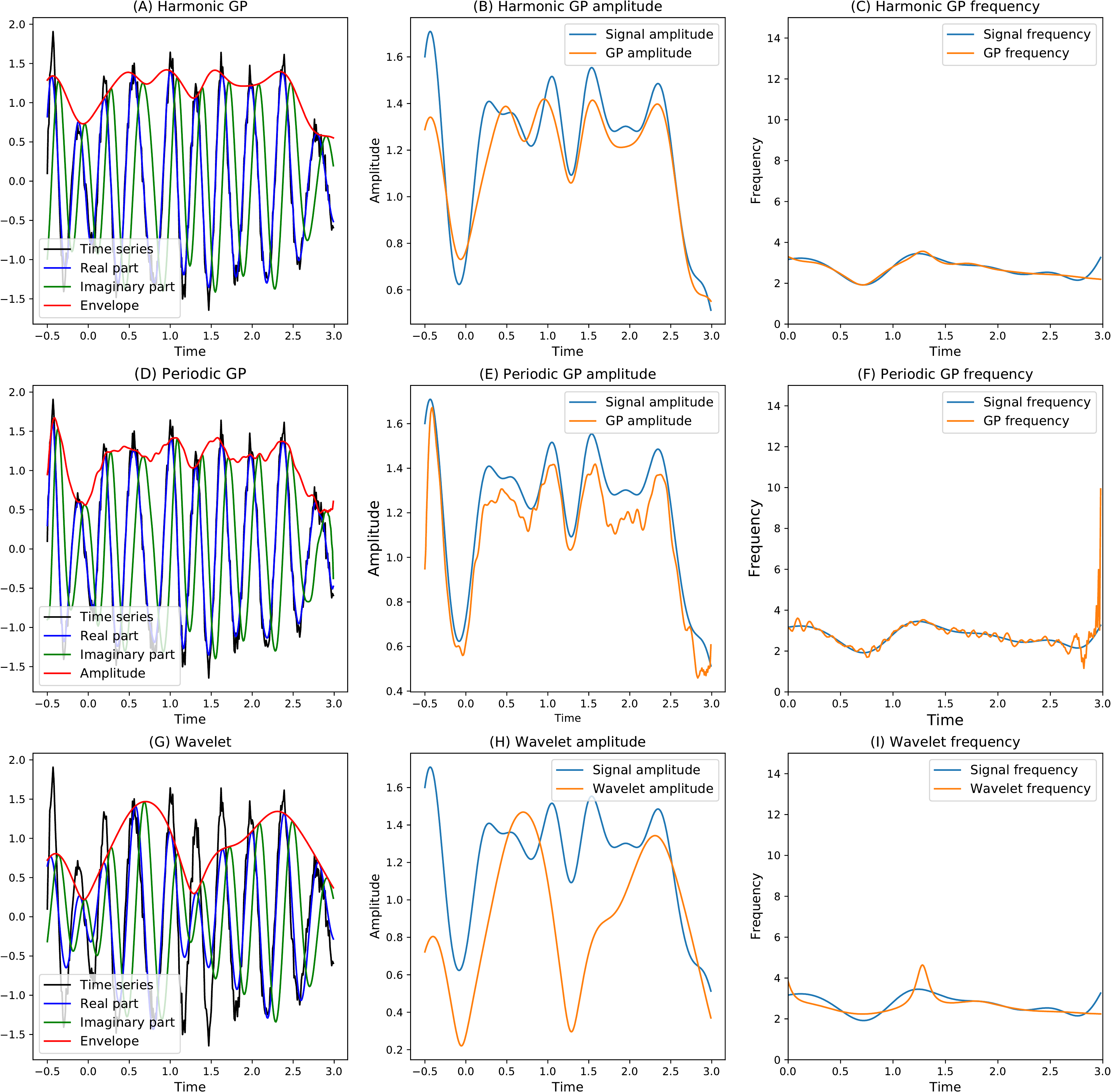}
	\caption{Analysis of a noise corrupted signal. A) Real part, imaginary part and amplitude of the harmonic CGPR complex signal. B,C) Instantaneous amplitude and frequency obtained through harmonic CGPR analysis. D) Real part, imaginary part and amplitude of the periodic CGPR complex signal. E,F) Instantaneous amplitude and frequency obtained through periodic CGPR analysis. G) Real part, imaginary part and amplitude of the wavelet signal. H,I) Instantaneous amplitude and frequency obtained using wavelet analysis.}
	\label{figure 7}
\end{figure}

\begin{figure}[!ht]
	\centering
    	\includegraphics[width=1.\textwidth] {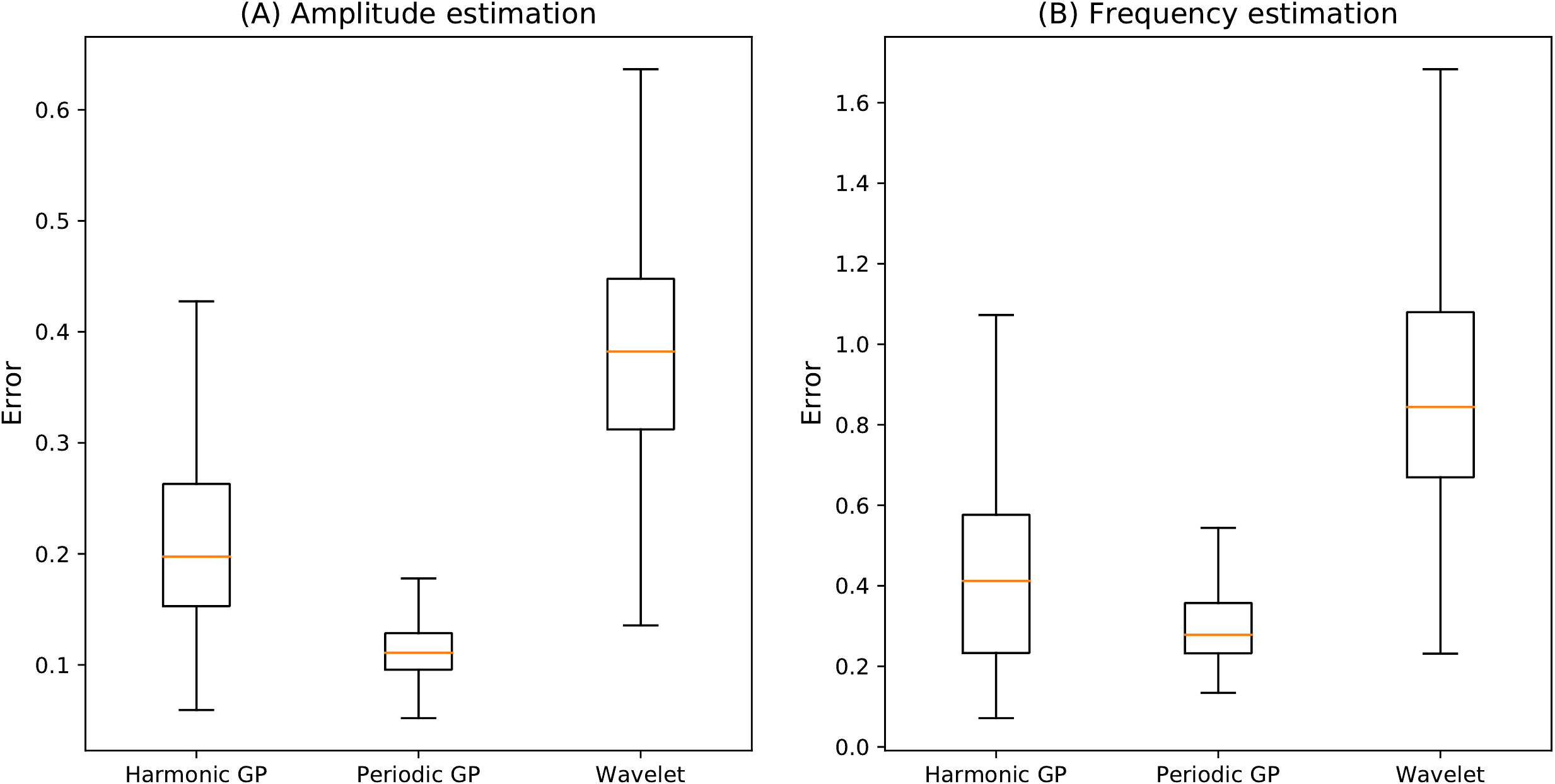}
	\caption{Comparative analysis of noise corrupted signals. (A,B) Comparison between harmonic CGPR, periodic CGPR and Morlet wavelet analysis.}
	\label{figure 8}
\end{figure}

Fig.~\ref{figure 7} shows the results of the analysis on an example signal. Fig.~\ref{figure 8} shows the error associated with each method. The periodic CGPR method outperforms both harmonic CGPR and Morlet wavelet analysis in both instantaneous amplitude and frequency estimation (p \textless~0.01). Similarly, the harmonic CGPR method performs significantly better than the wavelet analysis in both tasks (p\textless~0.01).


\subsection{Experimental results on brain data}
In this subsection we apply the CGPR to real MEG data and use it to study the non-stationary dynamics of brain waves. We focus on alpha oscillations (8-13 Hz). In this analysis we use the harmonic quasi-quadrature covariance function (Eq.~\ref{harmonic covariance, results}) since it is not destabilized by the non-harmonic waveform of alpha oscillations. 

\subsubsection{Data acquisition and preprocessing}
We collected resting state brain activity from an experimental participant that was instructed to fixate a cross at the center of a black screen. The study was conducted in 913
accordance with the Declaration of Helsinki and approved by the local ethics committee 914
(CMO Regio Arnhem-Nijmegen). Informed written consent was obtained from all 915
participants. Brain activity was recorded using a 275 axial gradiometer MEG setup (VSM/CTF Systems, Port Coquitlam, British Columbia, Canada). The acquisition sampling rate was 1200 Hz. Preprocessing was performed using the open source Matlab toolbox FieldTrip \citep{oostenveld2010fieldtrip}. The continuous data was cut in segments of two seconds. Trial segments containing movement, muscle, and SQUID artifacts were discarded following a semi-automatic artifact detection routine. A fourth-order Butterworth band-stop filter (49-51 Hz) was used to remove 50 Hz line noise. Finally, residual eye blinks, heartbeat, line noise and muscular artifacts were isolated by independent component analysis \citep{comon1994independent} on the concatenated segments and subsequently removed from the data. 

\subsection{Analysis of alpha oscillations}
\begin{figure}[!ht]
	\centering
    	\includegraphics[width=1.\textwidth] {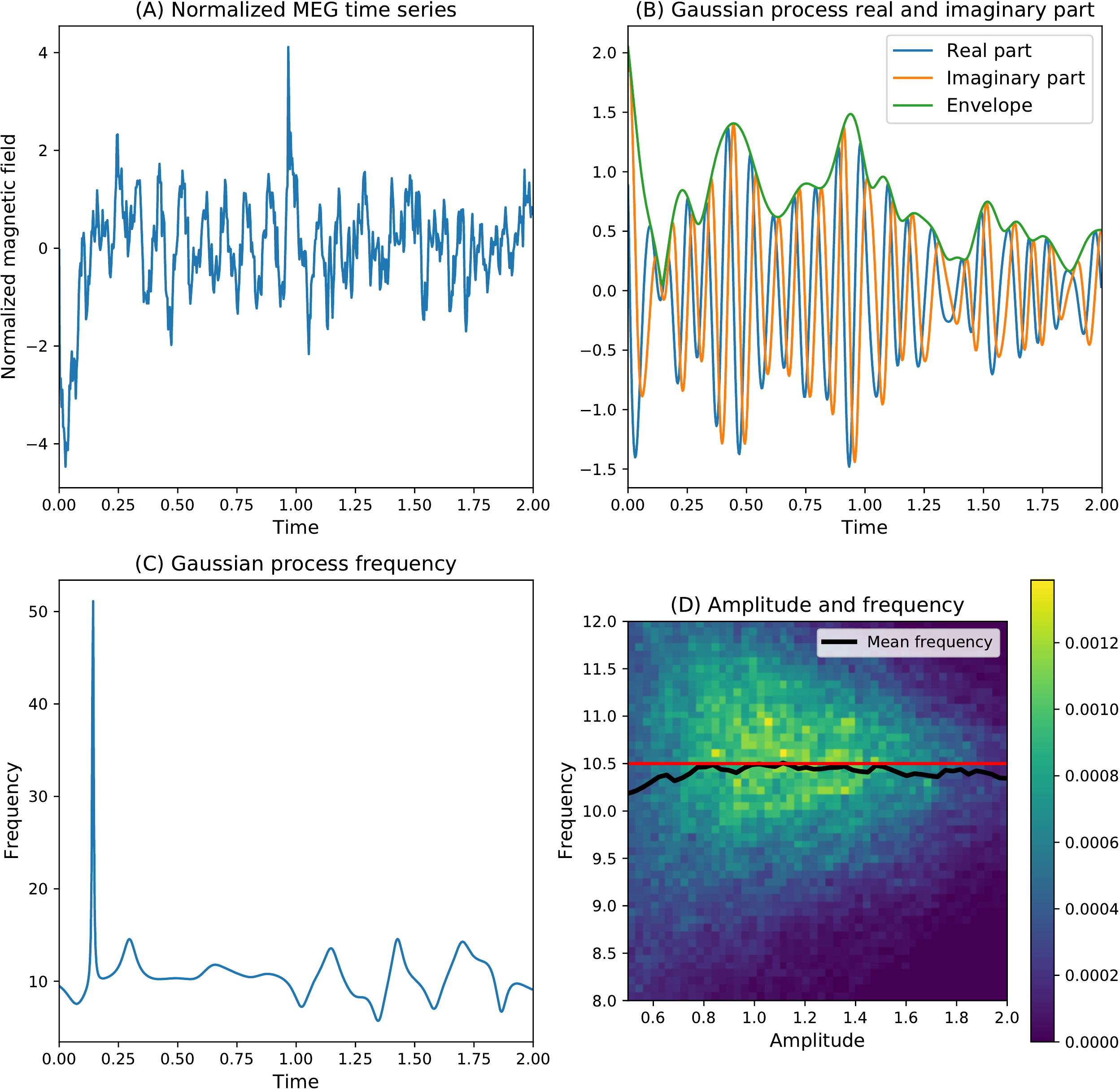}
	\caption{GP analysis of MEG data. A) Raw MEG signal of an example trial. B) Real and imaginary part of the GP-based complex signal obtained from the MEG signal. The real and the imaginary part are depicted in blue and green respectively. The amplitude is represented by the red line. C) Instantaneous frequency estimated using the CGPR. D) Density plot of instantaneous amplitude and frequency across all trials and time points. }
	\label{figure 9}
\end{figure}

We restricted our attention to the analysis of the MEG sensor with the highest alpha power, which was located in the occipital part of the helmet. We used the CGPR for constructing the real and the imaginary imaginary part of the oscillation and computing the instantaneous amplitude and frequency. We modeled the measurement noise as a superposition of Ornstein--Uhlenbeck and white noise:
\begin{equation}
Q_{j,k} = \nu^2 \exp{\big(- b |t_k - t_j|\big)} + \sigma^2 \mu_{k,j}
\label{noise covariance, results}
\end{equation}
We used this covariance matrix in Eq.~\ref{complex posterior expectation GP, methods}. The parameter $\sigma (= 0.5)$ is the standard deviation of the white noise while $\nu (= 1)$ and $b (= 1)$ are standard deviation, respectively, relaxation parameter (inverse of time constant) of the Ornstein--Uhlenbeck noise, here modeled using an exponential covariance function. The parameters of the noise and the oscillation were chosen in a way to fall within the range of real human MEG signals \citep{ambrogioni2016dynamic}. Fig.~\ref{figure 9}A shows an example MEG signal. The alpha oscillation is clearly visible in the initial and the final segment of the trial, although its amplitude visibly decreases in the middle. These fluctuations in the alpha amplitude are well known in the neuroscience community \citep{klimesch2011alpha}. Fig.~\ref{figure 9}B shows the real and the imaginary part of the GP-based complex signal obtained from this signal, together with its instantaneous amplitude. We see that the combination of harmonic quadrature covariance function and noise model given by Eq.~\ref{noise covariance, results} allows to effectively filter out the white and low frequency broadband noise and to obtain a smooth estimate of the amplitude envelope. 

We also investigated the relation between instantaneous amplitude and frequency. This analysis could potentially constrain the possible network mechanisms that could generate the alpha oscillation. Fig~\ref{figure 9}D shows the density plot of instantaneous amplitude and frequency across all trials and time points. The mean frequency does not seem to systematically change as a function of the amplitude, except for a small downward trend for very low and very high amplitudes. Conversely, the variance of the frequency clearly decreases when the instantaneous amplitude increases. The latter effect is most likely due to the relation between instantaneous amplitude and signal-to-noise ratio (SNR).

\section{Discussion}
We introduced a new method for estimating instantaneous amplitude, phase and frequency of narrow-band real-valued signals using a complex-valued version of GP regression. The main innovation of the paper is the definition of two new classes of complex-valued covariance functions (i.e., quadrature and quasi-quadrature covariance functions) that induce either an exact or an approximate quadrature relationship between the real and the imaginary part of random signals. Using these covariance functions, we can construct the unobservable imaginary part from the observable real part of the signal. A quadrature covariance function is obtained by applying the quadrature filter on a stationary real-valued covariance function and it defines GPs whose samples are analytic functions with probability one. We argued that, when the spectra of amplitude envelope and oscillation are highly overlapping, the analytic complex valued signals result in rather counter-intuitive estimates of instantaneous amplitude, phase and frequency. To obtain more plausible estimates for this type of signals, we introduced quasi-quadrature covariance functions. A quasi-quadrature covariance function is obtained by multiplying a quadrature covariance function with a real-valued envelope. We showed that these covariance functions specify GPs whose samples are not analytic while maintaining the quadrature relationships in expectation between the real and the imaginary part. 

By analyzing simulated chirplets and stochastic oscillations, we showed that the CGPR equipped with a quasi-quadrature covariance function outperforms the quadrature filter in recovering instantaneous amplitude and frequency. In particular, we found that the improvement is higher when the frequency of the signal was smaller. This result was expected as we showed that the deviation from the exact quadrature relationship between real and imaginary part is largest for very low frequencies as for these frequencies the spectra of the amplitude envelope and the oscillation overlap most. 

\subsection{Practical considerations}
We now discuss some practical considerations concerning the application of CGPR analysis on real data. An important feature of CGPR (and GP regression in general) is its ability to exploit prior knowledge about signal and noise processes through the choice of signal and noise covariance functions. For example, in our analysis of MEG data we modeled the signal as a complex-valued harmonic oscillation corrupted by two sources of real-valued noise: a white noise and an Ornstein-–Uhlenbeck process. Importantly, the parameters of the different covariance functions (both for the noise and the oscillatory processes) can be inferred from the signal itself. For example, the peak frequency parameter of the harmonic or the periodic covariance function can be optimized in order to increase the performance of the method. The optimization can be done by maximizing the marginal likelihood of the model \citep{rasmussen2006gaussian} or by a non-linear least-squares approach \citep{ambrogioni2016dynamic}. Since we assumed the imaginary part to not be directly observable, these optimization procedures are completely equivalent to the ones for the associated real-valued GP regression. 

One of the shortcomings of the CGPR is its cubic time complexity. In fact, the computation of the posterior expectation (Eq.~\ref{complex posterior expectation GP, methods}) requires the inversion of the prior covariance matrix between all sample points. The cubic complexity is problematic for the analysis of long signals. In our (non-optimized) script running on a single CPU, the analysis of $300$ trials (two seconds long) took 114 times longer than the Hilbert method (57 vs 0.5 seconds). Fortunately, there are several methods that can effectively deal with this problem. A possible solution is to use an envelope function with a compact support, i.e. one that is different from zero only for a short time lag. This kind of covariance functions allows to work with more computationally advantageous sparse covariance matrices because many of the entries are exactly zero \citep{wendland2004scattered}. An alternative approach is to convert the CGPR into a complex-valued infinite dimensional Kalman filter that can be solved in linear time \citep{sarkka2013spatiotemporal, solin2014explicit}. This conversion is exact for a large class of stationary covariance functions and approximate but arbitrarily accurate for the others. Another approach is to perform the CGPR in the frequency domain, which gives linear time complexity but introduces periodic boundary conditions as it requires a prior Fourier transform of the data \citep{paciorek2007bayesian}.

\subsection{Extensions of the CGPR method}
The CGPR analysis can be adapted to non-Gaussian data by adopting different likelihood models. For example, the analysis can be applied to binary data by introducing a latent complex-valued Gaussian process whose real part determines the probability of a Bernoulli variable \citep{rasmussen2006gaussian}. This approach can be used to estimate the instantaneous phase and frequency of a binary signal. Using a similar approach (adopting a different likelihood) can also be used to make the analysis can also be made more resistant to outliers. Specifically, this can be achieved by using a student-t instead of a Gaussian likelihood \citep{jylanki2011robust}. 

\subsection{Conclusion}
In sum, we have described a new method for estimating instantaneous amplitude, phase and frquence of narrow-band real-valued signals using a complex-valued version of GP regression (CGPR). We showed that CGPR with quasi-quadrature covariance functions provides a much better estimate of the instantaneous amplitude and frequency than the quadrature filter and Morlet wavelet analysis. CGPR is a versatile tool because it allows to incorporate prior information about the structure in signal and noise and thereby to tailor the analysis to the features of the signal.

\section{Appendix I: Hermitianity and positive semi-definiteness of quadrature and quasi-covariance functions}

In this appendix we show that quadrature and quasi-quadrature covariance functions are both positive semi-definite and Hermitian. A kernel function $\mathfrak{K}(\tau)$ is said to be Hermitian when:
\begin{equation}
\mathfrak{K}(\tau) = \mathfrak{K}(-\tau)^{*}~.
\label{hermitian first definition, appendix I}
\end{equation}
This implies that the real part of $\mathfrak{K}(\tau)$ is a even function while its imaginary part is an odd function. The quadrature covariance function $\mathcal{A} k(\tau)$ is obtained by applying the quadrature filter $\mathcal{A}$ on a stationary covariance function $k(\tau)$. On one hand, the real part is $\mathcal{A} k(\tau)$ is equal to $k(\tau)$ and is therefore symmetric (otherwise $k(\tau)$ would not be a valid covariance function). On the other hand, the imaginary part $\mathcal{H} k(\tau)$ is odd since 
\begin{equation}
\mathcal{H}k(-\tau) = \int_{-\infty}^{+\infty} \frac{k(s)}{-\tau - s} ds = -\int_{-\infty}^{+\infty} \frac{k(r)}{\tau - r} dr = -\mathcal{H}k(\tau)
\label{odd imaginary part, appendix I}
\end{equation}
where we defined the new integration variable $r = - s$ and leveraged the fact that $k(\tau)$ is even. The Hermitianity implies that the Fourier transform is real-valued. This follow from
\begin{equation}
\mathcal{F}[\mathcal{A}k(\tau)](\xi) = \mathcal{F}[k(\tau)](\xi) + i \mathcal{F}[\mathcal{H}k(\tau)](\xi)~.
\label{real spectrum, appendix I}
\end{equation}
In fact, the first Fourier transform on the right hand side is real-valued since $k(\tau)$ is even while the second Fourier transform is purely imaginary since $k(\tau)$ is odd.

An Hermitian kernel function $\mathfrak{K}(\tau)$ is said to be positive semi-definite when its Fourier transform is non-negative (almost) everywhere. The Fourier transform $\mathcal{A} k(\tau)$ is $2 h(\xi) \tilde{k}(\xi)$ that is obviously non-negative if $k(\tau)$ is a valid covariance function.

Quasi-quadrature covariance functions are obtained by multiplying a quadrature covariance function with a real-valued even positive definite envelope function. Clearly the pointwise product of an Hermitian function with a real-valued even function is itself Hermitian. Furthermore, a pointwise product of two positive semi-definite functions is always positive semi-definite because the convolution of two non-negative functions is always non-negative.  Therefore, quasi-quadrature covariance functions are indeed valid covariance functions.

\section{Appendix II: Derivation of cross-covariance matrix, cross-spectral density and coherency}
Here, we derive formula for the entries of the cross-covariance matrix of $\Psi(t)$. This matrix is defined by the following formula
\begin{equation}
\Theta(\tau) = \langle \Psi(t) \Psi(t)^T \rangle = 
\begin{pmatrix}
    \langle \alpha(t) \alpha(t + \tau) \rangle & \langle \alpha(t) \beta(t + \tau) \rangle \\
    \langle \beta(t) \alpha(t + \tau) \rangle  & \langle \beta(t) \beta(t + \tau) \rangle
\end{pmatrix}.
\label{cross-covariance function, appendix II}
\end{equation}
The autocovariance of the real part can be obtained by rewriting $\alpha(t)$ as $\frac{1}{2} \big( \zeta(t) + \zeta(t)^* \big)$. In fact, by plugging this formula on Eq.~\ref{cross-covariance function, appendix II} we obtain
\begin{equation}
\langle \alpha(t) \alpha(t + \tau) \rangle = \frac{1}{4}\bigg( \langle \zeta(t) \zeta(t + \tau)\rangle + \langle \zeta(t) \zeta(t + \tau)^*\rangle + \langle \zeta(t)^* \zeta(t + \tau)\rangle + \langle \zeta(t)^* \zeta(t + \tau)^*\rangle \bigg) 
\label{auto-covariance derivation I, appendix II}
\end{equation}
The pseudo auto-covariance functions $\langle \zeta(t) \zeta(t + \tau)\rangle$ and $\langle \zeta(t)^* \zeta(t + \tau)^*\rangle$ vanish since $\zeta(t)$ is circularly-symmetric. Furthermore $$\frac{1}{4} \bigg( \langle \zeta(t) \zeta(t + \tau)^*\rangle + \langle \zeta(t)^* \zeta(t + \tau)\rangle \bigg) = \frac{1}{4}\bigg( \mathcal{A}k(\tau) + {\mathcal{A}k(\tau)}^* \bigg) = \frac{1}{2} \mathfrak{R}\mathcal{A}k(\tau) = \frac{1}{2} k(\tau).$$
By an analogous reasoning, it can be shown that $\langle \beta(t) \beta(t + \tau) \rangle = \frac{1}{2} k(\tau)$, $\langle \alpha(t) \beta(t + \tau) \rangle = - \frac{1}{2} \mathcal{H}k(\tau)$ and $\langle \beta(t) \alpha(t + \tau) \rangle = \frac{1}{2} \mathcal{H}k(\tau)$.

The cross-spectral density function is defined as the Fourier transform of the cross-covariance function. We can write its entries as follow:
\begin{equation}
\tilde{\Theta}(\xi) = 
\begin{pmatrix}
    \tilde{\Theta}(\xi)_{11} & \tilde{\Theta}(\xi)_{12} \\
    \tilde{\Theta}(\xi)_{21}  & \tilde{\Theta}(\xi)_{22}
\end{pmatrix}.
\label{cross-spectral function, appendix II}
\end{equation}
The coherency $\gamma(\xi)$ is a complex number that can be obtained from the cross-spectral density function as follow:
\begin{equation}
\gamma(\xi) = \frac{\tilde{\Theta}(\xi)_{12}}{\sqrt{\tilde{\Theta}(\xi)_{11} \tilde{\Theta}(\xi)_{22}}}
\label{coherency, appendix II}
\end{equation}
The matrices in Eq.~\ref{cross-covariance function, appendix II} can always be diagonalized in the following way: 
\begin{equation}
\tilde{\Theta}(\xi) = 
\begin{pmatrix}
    v^{(1)}_1(\xi) & v^{(2)}_1(\xi) \\
    v^{(1)}_2(\xi)  & v^{(2)}_2(\xi)
\end{pmatrix}
\begin{pmatrix}
    \lambda^{(1)}(\xi) & 0 \\
    0  & \lambda^{(2)}(\xi)
\end{pmatrix}
\begin{pmatrix}
    v^{(1)}_1(\xi) & v^{(1)}_2(\xi) \\
    v^{(2)}_1(\xi)  & v^{(2)}_2(\xi)
\end{pmatrix}~.
\label{diagonalized cross-covariance function, appendix II}
\end{equation}
where the eigenvalues $\lambda^{(1)}(\xi)$ and $\lambda^{(2)}(\xi)$ are non-negative real numbers. By computing the entries of this matrix product, we can express the coherency in terms of the eigenvalues and eigenvectors
\begin{equation}
\gamma(\xi) = \frac{v^{(1)}_1(\xi) {v^{(1)}_2(\xi)}^* \lambda^{(1)}(\xi) + v^{(2)}_1(\xi) {v^{(2)}_2(\xi)}^* \lambda^{(2)}(\xi)}{\sqrt{\big( |v^{(1)}_1|^2 \lambda^{(1)}(\xi) + |v^{(2)}_1|^2 \lambda^{(2)}(\xi) \big) \big( |v^{(1)}_2|^2 \lambda^{(1)}(\xi) + |v^{(2)}_2|^2 \lambda^{(2)}(\xi) \big) }}~.
\label{coherency from eigendecomposition, appendix II}
\end{equation}

\section{Appendix III: Proof of the asymptotic dominance of the first eigenvalue}
In this appendix, we prove that Eq.~\ref{envelope requirements II, method} implies 
$$
\lim_{\xi\to\infty} \frac{\lambda_f^{(2)}(\xi)}{\lambda_f^{(1)}(\xi)} = 0~.
$$
From Eq.~\ref{eigenvalue quasi-quadrature, methods}, we have 
$$
\frac{\lambda_f^{(2)}(\xi)}{\lambda_f^{(1)}(\xi)} = \frac{\int_{-\infty}^{0} \tilde{f}(\xi - \upsilon) \tilde{k}(\upsilon) d\upsilon}{\int_{0}^{+\infty} \tilde{f}(\xi - \upsilon) \tilde{k}(\upsilon) d\upsilon}~.
$$
From Eq.~\ref{envelope requirements II, method} and the fact that $\tilde{k}(\upsilon)$ is positive-valued, we have that there exist a positive number $\xi^*$ such that, for $\forall \xi > \xi^*$
$$
\frac{\int_{-\infty}^{0} \tilde{f}(\xi - \upsilon) \tilde{k}(\upsilon) d\upsilon}{\int_{0}^{+\infty} \tilde{f}(\xi - \upsilon) \tilde{k}(\upsilon) d\upsilon} < \frac{\big(\int_{-\infty}^{0} \tilde{k}(\upsilon) d\upsilon \big) f(\xi)}{\int_{0}^{+\infty} \tilde{f}(\xi - \upsilon) \tilde{k}(\upsilon) d\upsilon} = \frac{\big(\int_{-\infty}^{0} \tilde{k}(\upsilon) d\upsilon \big)}{\int_{0}^{+\infty} \frac{\tilde{f}(\xi - \upsilon)}{f(\xi)} \tilde{k}(\upsilon) d\upsilon}~.
$$
The integral in the denominator of the rightmost expression diverges as $\xi$ tends to infinity. This follows from Eq.~\ref{envelope requirements II, method} and from the positivity of $\tilde{k}(\upsilon)$. Hence
$$
\lim_{\xi\to\infty}  \frac{\big(\int_{-\infty}^{0} \tilde{k}(\upsilon) d\upsilon \big)}{\int_{0}^{+\infty} \frac{\tilde{f}(\xi - \upsilon)}{f(\xi - \upsilon^*)} \tilde{k}(\upsilon) d\upsilon} = 0~.
$$
This concludes the proof since, for a sufficently large value of $\xi$, this expression bounds $\lambda_f^{(2)}(\xi)/\lambda_f^{(1)}(\xi)$ from above.
\bibliography{complexGPbiblio}

\end{document}